\documentclass[10pt,twocolumn,letterpaper]{article}

\usepackage[pagenumbers]{cvpr} 

\usepackage{multirow}
\usepackage{graphicx}
\usepackage[table,xcdraw]{xcolor}
\usepackage{capt-of}
\usepackage[most]{tcolorbox} 
\usepackage{enumitem}
\usepackage{colortbl}

\usepackage{algorithm}
\usepackage{algorithmic}

\newtcolorbox{promptbox}[2][]{%
    enhanced,
    attach boxed title to top left={yshift=-3mm, xshift=3mm},
    colback=gray!5!white,
    colframe=gray!60!black,
    colbacktitle=gray!20!white,
    coltitle=black,
    fonttitle=\bfseries\small,
    title={#2},
    boxrule=0.5pt,
    arc=2mm,
    left=2mm, right=2mm, top=3mm, bottom=2mm, 
    fontupper=\footnotesize\linespread{1.05}\selectfont, 
    #1
}

\definecolor{cvprblue}{rgb}{0.21,0.49,0.74}
\usepackage[pagebackref,breaklinks,colorlinks,allcolors=cvprblue]{hyperref}

\def\paperID{} 
\def\confName{}
\def\confYear{}

\title{Asking like Socrates: Socrates helps VLMs understand remote sensing images}

\author{
    Run Shao$^{1,2,*}$\quad 
    Ziyu Li$^{1,*}$\quad 
    Zhaoyang Zhang$^{1}$\quad 
    Linrui Xu$^{1}$\quad 
    Xinran He$^2$\quad
    Hongyuan Yuan$^{1,2}$\and
    Bolei He$^2$\quad 
    Yongxing Dai$^2$\quad 
    Yiming Yan$^3$\quad 
    Yijun Chen$^3$\quad 
    Wang Guo$^1$\quad
    Haifeng Li$^{1,\dagger}$\\ \\
    $^1$School of Geosciences and Info-Physics, Central South University, Changsha, China\\
    $^2$Baidu Inc., Beijing, China\\
    $^3$School of Earth Sciences, Zhejiang University, Hangzhou, China\\
}

\newcommand{\OurModelName}{RS-EoT-7B}
\newcommand{\OurSFTData}{RS-EoT-4K}

\begin{document}
\twocolumn[{%
\maketitle
\begin{center}
    \centering
    \includegraphics[width=\linewidth]{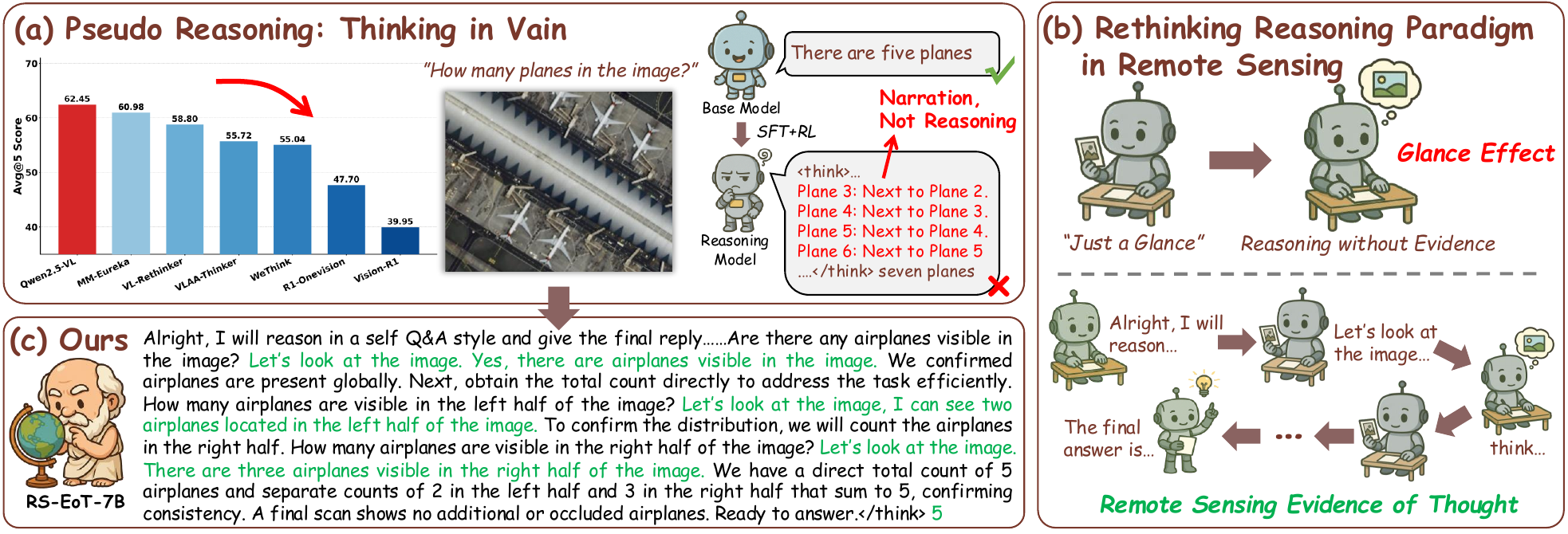}
    \vspace{-20pt}
    \captionof{figure}{Illustration of the pseudo reasoning problem and our RS-EoT solution. (a) Existing models show pseudo reasoning: explicit thinking (blue bars) degrades performance below the non-reasoning base model (red bar). (b) We attribute this to the ``Glance Effect''—reasoning based on a single, coarse perception. We propose RS-EoT, an iterative evidence-seeking loop. (c) Our model, RS-EoT-7B, successfully solves the task by iteratively reasoning and seeking visual evidence.}
  \label{fig:overview}
\end{center}
}]

\renewcommand{\thefootnote}{\fnsymbol{footnote}} 
\footnotetext[1]{Equal contribution.}
\footnotetext[2]{Corresponding author.}

\begin{abstract}
Recent multimodal reasoning models, inspired by DeepSeek-R1, have significantly advanced vision–language systems. However, in remote sensing (RS) tasks, we observe widespread pseudo reasoning: models narrate the process of reasoning rather than genuinely reason toward the correct answer based on visual evidence. We attribute this to the Glance Effect, where a single, coarse perception of large-scale RS imagery results in incomplete understanding and reasoning based on linguistic self-consistency instead of visual evidence. To address this, we propose RS-EoT (Remote Sensing Evidence-of-Thought), a language-driven, iterative visual evidence-seeking paradigm. To instill this paradigm, we propose SocraticAgent, a self-play multi-agent system that synthesizes reasoning traces via alternating cycles of reasoning and visual inspection. To enhance and generalize these patterns, we propose a two-stage progressive RL strategy: first, RL on fine-grained Grounding tasks to enhance RS-EoT capabilities, followed by RL on RS VQA to generalize to broader understanding scenarios. Experiments show RS-EoT achieves state-of-the-art performance on multiple RS VQA and grounding benchmarks. Analyses reveal clear iterative cycles of reasoning and evidence seeking, confirming RS-EoT mitigates the Glance Effect and enables genuine evidence-grounded reasoning. Our code, data, and models are available at \url{https://geox-lab.github.io/Asking_like_Socrates}.
\end{abstract}    
\section{Introduction}
\label{sec:intro}

Recent deep reasoning models, such as DeepSeek-R1~\cite{deepseek-r1}, Qwen3~\cite{yang2025qwen3}, and Doubao-Seed-1.6~\cite{doubao-seed-1.6}, show the SFT–RL paradigm substantially enhances long-chain reasoning~\cite{wen-etal-2025-light,team2025kimi,huang2025vision,meng2025mm}, enabling breakthroughs in math, code, and agent planning. This paradigm was then extended to multimodal settings, training vision–language models to generate structured reasoning~\cite{huang2025vision,meng2025mm,yang2025wethink,chen2025sft,wang2025vl,yang2025r1}. This progress has significantly advanced multimodal systems, stimulating interest in applying deep reasoning to complex visual domains~\cite{yuan2025vl,saxena2025uav,zhang2025grounded,tao2025advancements,liu2025remote}.

However, owing to the unique properties of remote sensing (RS) imagery, such as wide spatial extents, large scale variations, and sparse and subtle visual clues, these models often exhibit \emph{pseudo reasoning}~\cite{tao2025advancements,li2025ddfav} in RS tasks (\cref{fig:overview}-a). Despite producing explicit reasoning chains, their performance on RS tasks shows no gain~\cite{chen2025sft,shen2025satori,fang2025thinkless,fan2025missing}, and even degrades compared to the base model. In other words, they merely \emph{describe} a reasoning process rather than \emph{performing} one. We attribute this to the \emph{Glance Effect} (\cref{fig:overview}-b). Current models perform only a single, coarse perception pass (a ``glance'') before reasoning. This is insufficient, leading to reasoning built on incomplete visual evidence. Consequently, the model drifts toward linguistically self-consistent narratives rather than evidence-based logic, which can even degrade performance.

This suggests RS reasoning requires iterative, not static, evidence acquisition~\cite{tao2025advancements,vlm_rs}. Human analysts exemplify this, using repeated cycles of inspection and refinement. To emulate this, models must adopt a paradigm where reasoning guides perception to dynamically seek new visual evidence, rather than relying on a fixed initial view.

To this end, we propose \textit{RS-EoT (Remote Sensing Evidence-of-Thought)}, a language-driven iterative evidence-seeking reasoning paradigm. It frames reasoning as a reasoning–perception loop where the model continuously revisits the image, seeking new visual cues guided by the evolving reasoning. Thus, perceptual grounding is progressively refined, not fixed at the outset. To instill RS-EoT, we follow Deepseek-R1's methodology: first, Supervised Fine-Tuning (SFT) cold-starts the reasoning mode, and then Reinforcement Learning (RL) enhances and generalizes these patterns.~\cite{deepseek-r1,huang2025vision,meng2025mm,yang2025r1}.

In the SFT phase, we propose a novel data synthesis method called SocraticAgent to equip the model with the aforementioned self-iterative evidence-seeking capability. Inspired by the Socratic Method, which fosters knowledge acquisition through a step-by-step questioning process rather than direct instruction, we design SocraticAgent as a self-play multi-agent system to synthesize RS-EoT reasoning traces. Specifically, SocraticAgent contains a Reasoner and a Perceiver. The former is responsible for pure-language reasoning and posing perceptual questions about the image as necessitated by the reasoning, while the latter is responsible for perceiving the image and answering the questions posed by the Reasoner. They engage in a multi-turn dialogue, simulating the iterative reasoning-perception loop. Through a Socratic self-play strategy, wherein each agent is prompted with the presumed limited capability of the other, the Reasoner is guided to pose simple, incremental questions and the Perceiver to return accurate yet concise results, ensuring detailed and progressive traces. This synthesis method yielded the RS-EoT-4K dataset, which is used for the initial SFT cold-start to instill this self-iterative evidence-seeking reasoning in the model. 


In the RL stage, we propose a two-stage progressive training pipeline to further enhance and generalize the RS-EoT capability. First, adhering to the principle that ``iron sharpens iron,'' we perform RL on fine-grained grounding tasks, which are precisely those that most demand fine-grained visual evidence, to specifically enhance the RS-EoT capability. Building on this foundation, we then conduct RL on general RS VQA tasks to generalize this capability to broader remote sensing scenarios. However, considering that existing RS VQA data is often simple (\eg, Yes/No types), it is highly susceptible to reward hacking and thus unsuitable for RL training. To address this, we propose a multiple-choice VQA reconstruction strategy and a tailored reward function, enabling stable RL training.

Experiments show RS-EoT achieves state-of-the-art performance on multiple RS VQA and grounding benchmarks. Case studies and attention analyses confirm the model employs clear iterative cycles of reasoning and evidence seeking, validating the effectiveness of RS-EoT.

Our main contributions are summarized as follows:

\begin{itemize}[leftmargin=1.4em]
\item We propose \textbf{RS-EoT} (Remote Sensing Evidence-of-Thought), a language-driven iterative evidence-seeking reasoning paradigm for remote sensing.
\item We introduce a complete framework to instill this paradigm: (1) \textbf{SocraticAgent} to synthesize the \textbf{RS-EoT-4K} dataset for SFT cold-starting. (2) A two-stage \textbf{progressive Grounding+VQA RL} strategy, featuring a novel VQA data reconstruction method that enables stable RL training on simple RS VQA datasets.
\item We train \textbf{RS-EoT-7B}, which achieves SOTA on multiple RS VQA and grounding benchmarks. Analyses confirm it mitigates the Glance Effect via clear iterative reasoning and evidence-seeking cycles.
\end{itemize}
\section{Related Work}
\label{sec:relatedwork}

\subsection{Vision-language Model for Remote Sensing}
\label{sec:rw_vlm_rs}

The automated interpretation of RS images is crucial for applications like environmental monitoring, urban planning, and disaster response~\cite{vlm_rs,huang2025survey_rs_foundationmodel}. This has spurred development of VLMs for RS, with models like GeoChat~\cite{geochat}, Skysense~\cite{guo2024skysense}, and RingmoGPT~\cite{wang2024ringmogpt} demonstrating strong image understanding capabilities. More recently, the field has advanced toward complex reasoning, with models like Geo-R1~\cite{geo-r1} and VHM-RL~\cite{vhm-rl} adopting the SFT-RL paradigm to output explicit reasoning chains.

However, a key issue persists: these models often rely on a single-pass, global perception, neglecting detailed information and regional cues. This is particularly problematic for RS images with their wide geographical ranges and large scale variations. In contrast, our work introduces RS-EoT, an iterative evidence-seeking paradigm that replaces this single-pass approach with a dynamic loop of linguistic deduction and targeted visual inspection.

\subsection{Large Language Reasoning Model}
\label{sec:rw_reasoning_model}



Reasoning in LLMs was reshaped by the Chain-of-Thought (CoT) paradigm~\cite{cot}, which showed that eliciting multi-step traces improves complex reasoning. This inspired variants like Least-to-Most~\cite{Zhou2022LeasttoMostPE}, Program-of-Thought~\cite{chen2022program}, and advanced models like OpenAI’s o1~\cite{jaech2024openai}. A pivotal development is the SFT–RL paradigm, popularized by DeepSeek-R1~\cite{deepseek-r1}. This approach uses SFT to cold-start reasoning patterns, then applies RL to refine and generalize them. Algorithms like Group Relative Policy Optimization (GRPO)~\cite{shao2024deepseekmath} and its variants~\cite{wen-etal-2025-light,yu2025dapo} offer more stable optimization for long-sequence reasoning than PPO~\cite{ppo}. Our methodology builds upon this SFT-RL paradigm: we first employ SFT to cold-start the iterative evidence-seeking pattern, then utilize a two-stage progressive RL process to enhance and generalize this capability.

\subsection{Multimodal reasoning model}
\label{sec:rw_cot_model}


The SFT–RL paradigm was rapidly adopted in multimodal learning, spawning models like Vision-R1~\cite{huang2025vision}, WeThink~\cite{yang2025wethink}, and R1-OneVision~\cite{yang2025r1}. These models, with subsequent refinements~\cite{chen2025sft,wang2025vl,meng2025mm}, transferred this methodology to visual-language contexts. However, they largely inherit language-oriented patterns, assuming reasoning can occur over a static, global visual representation. This assumption fails in remote sensing, where tasks demand iterative, region-specific inspection.

To address this gap, we propose RS-EoT, an iterative, evidence-seeking paradigm tailored for remote sensing. We introduce a complete framework to instill this capability, featuring: (1) a Socratic-inspired mechanism to synthesize RS-specific reasoning traces, and (2) a multi-stage progressive training pipeline to instill this behavior. This design overcomes the limitations of single-pass, language-oriented reasoning frameworks.
\section{Method}
\label{sec:method}

\begin{figure*}[t]
  \centering
  \includegraphics[width=\linewidth]{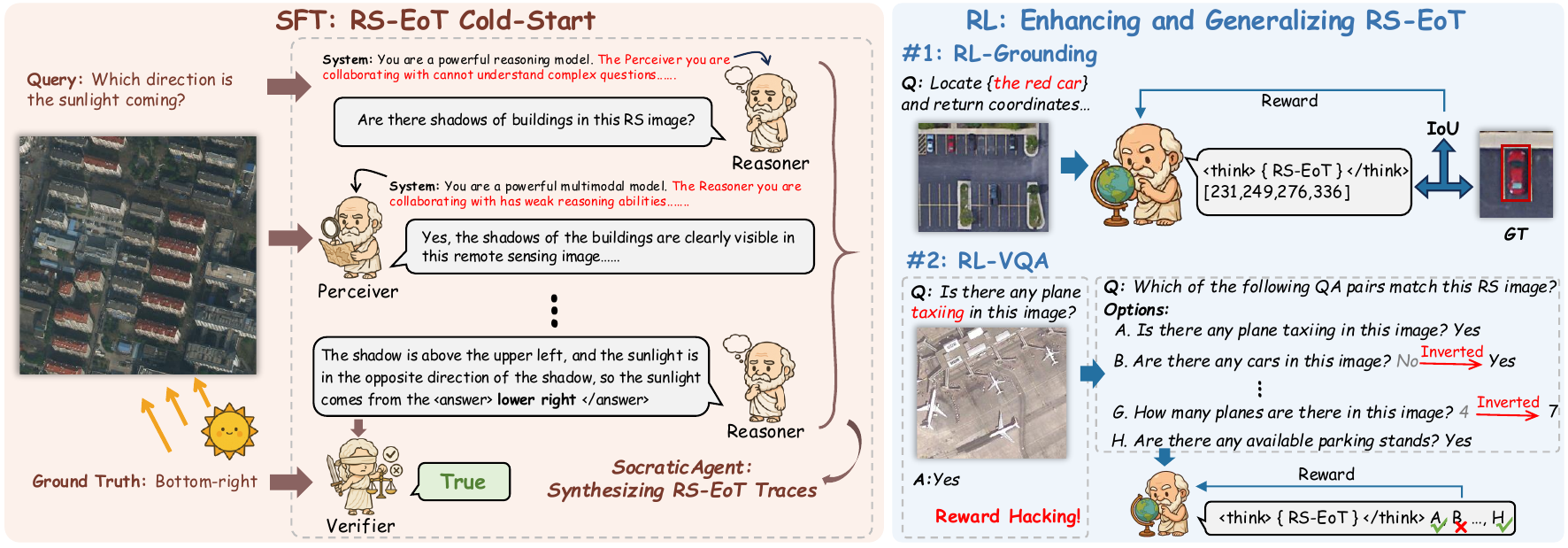}
  \vspace{-15pt}
  \caption{Overview of our method to instill the RS-EoT paradigm. \textbf{(Left) SFT: RS-EoT Cold-Start:} We propose SocraticAgent to synthesize reasoning traces. A Reasoner (text-only) and a Perceiver (image-aware) engage in an iterative dialogue, guided by a self-play prompting mechanism. \textbf{(Right) RL: Enhancing and Generalizing RS-EoT:} A two-stage progressive RL pipeline. Stage 1 (RL-Grounding) enhances fine-grained evidence-seeking via an IoU-based reward. Building on this, Stage 2 (RL-VQA) generalizes reasoning by converting simple VQA datasets into a multiple-choice format with a graded reward for stable training.}
  \vspace{-5pt}
  \label{fig:method_overview}
\end{figure*}

We aim to enable multimodal models to perform evidence-seeking reasoning over remote sensing imagery. To this end, we first introduce \textbf{RS-EoT (Remote Sensing Evidence-of-Thought)}, which characterizes the desired reasoning pattern for remote sensing: reasoning is language-driven, iteratively seeking visual evidence to refine the reasoning chain and ultimately converge on the correct answer. Following the Deepseek-R1, we first inject the RS-EoT reasoning mode via SFT cold-starting, and then enhance and generalize it through RL. An overview of our method is presented in \cref{fig:method_overview}.

To instill this reasoning pattern, a primary challenge is synthesizing the cold-start data. To this end, we propose \textbf{SocraticAgent}, a self-play multi-agent system that synthesizes RS-EoT reasoning traces. SocraticAgent simulates an iterative process of reasoning (via a Reasoner) and visual evidence seeking (via a Perceiver), ensuring each reasoning step is grounded in verifiable visual evidence. These synthesized traces are then used for SFT to cold-start the RS-EoT reasoning mode.

Building on this SFT foundation, we adopt a \textbf{two-stage progressive RL} strategy to further enhance and generalize this capability. The first stage performs RL on fine-grained grounding tasks. Since these tasks naturally demand iterative evidence seeking, this stage effectively strengthens the RS-EoT behavior learned during SFT. The second stage performs RL on general remote sensing VQA, allowing the learned RS-EoT pattern to generalize to broader RS understanding scenarios. The final model obtained from this process is termed \textbf{\OurModelName{}}.

\subsection{SFT: RS-EoT Cold-Start}
\label{sec:socraticagent}

\subsubsection{RS-EoT}
\label{sec:rs_eot}

The RS-EoT reasoning paradigm is defined by two core principles. First, \textbf{reasoning is orchestrated by natural language}, which serves as the primary medium for formulating hypotheses, planning evidence-seeking steps, and refining intermediate conclusions. Language is not merely a description of multimodal information but the active controller that determines when and how to probe for multimodal evidence. Second, \textbf{visual information functions as on-demand evidence within the reasoning chain}. Instead of relying on a single, static global perception, the model actively seeks, verifies, and integrates localized visual evidence at each reasoning step. Through this iterative loop of language-driven reasoning and targeted evidence retrieval, RS-EoT enables more faithful, interpretable, and evidence-grounded decision-making in remote sensing tasks.

\subsubsection{Asking like Socrates}
\label{sec:asking_like_socrates}

A core challenge is \textit{how to synthesize RS-EoT reasoning traces from scratch}, given that existing models do not possess this reasoning mode. We draw inspiration from the Socratic method, which encourages knowledge discovery through a structured process of questioning rather than direct instruction, proposing that the iterative process of reasoning and evidence seeking can be framed as a continuous, incremental process of inquiry.

For instance, when addressing the problem, ``What is the heading of the ship in this remote sensing image?'' one might first naturally ask, ``Is there a ship in the image?'' This is followed by a coarse visual inspection to confirm its presence. Next, one would reason based on common sense: if the ship is moving, its wake might indicate the direction of travel. This hypothesis, derived from reasoning, leads to a new, more specific question: ``In which direction is the ship's wake pointing?'' This prompts a detailed visual inspection, and by observing the wake, one can finally infer the ship's heading (as the opposite direction of the wake).

By emulating this Socratic questioning, an iterative process of posing questions, seeking visual evidence, and refining the reasoning chain, we can progressively correct and complete our logical path to correctly solve the problem. Simulating the aforementioned process, we propose SocraticAgent, a self-play multi-agent system capable of synthesizing RS-EoT reasoning traces from remote sensing VQA datasets. It primarily comprises two roles: the Reasoner and the Perceiver. 

\begin{itemize}[leftmargin=1.4em] 

\item \textbf{Reasoner.} Instantiated by GPT-5-mini~\cite{openai-gpt5-systemcard}, the Reasoner operates solely on textual queries and basic image metadata (\eg, modality, count, size), lacking direct access to the RS images. It performs pure-text reasoning and is prompted to pose perceptual questions to the Perceiver when visual information is needed. Upon receiving the Perceiver's answers, it continues to reason or pose further questions until it can formulate a final answer. Throughout the interaction, the Reasoner continuously analyzes, summarizes, and integrates the Perceiver's feedback to build a coherent semantic understanding, ultimately producing the final answer.

\item \textbf{Perceiver.} Implemented using a powerful multimodal model (Gemini-2.5-flash~\cite{comanici2025gemini} in our setup), the Perceiver's input consists solely of the RS image and the questions posed by the Reasoner, without access to the original task query. Its sole task is to receive the Reasoner's questions and provide accurate answers based on the remote sensing image.

\end{itemize}

They engage in a multi-turn dialogue, culminating in the Reasoner providing the final answer. We then introduce a \textbf{Verifier} (instantiated by doubao-seed-1.6-thinking~\cite{doubao-seed-1.6}) to validate the Reasoner's answer against the ground truth. The design rationale is: if the Reasoner, despite having no direct visual access, arrives at the correct solution, the intermediate dialogue is deemed a reliable reasoning trace. This dialogue history is then concatenated based on the following template and formatted into a self QA style reasoning trajectory. \cref{fig:case_study} illustrates an actual case.

\begin{tcolorbox}[enhanced,
  colback=white,colframe=black!20,boxrule=0.4pt,arc=1pt,
  left=5pt,right=5pt,top=2pt,bottom=2pt,
  title={SocraticAgent EoT Trace Concatenation Template}, fonttitle=\bfseries,
  fontupper=\ttfamily\footnotesize]
\setlength{\parskip}{1pt}\setlength{\parindent}{0pt}
\textless think\textgreater\ Alright, I will reason in a self QA style and give the final reply.\par
\{Reasoner (round $1$): thought \& question\}\\[-0pt]
\{Perceiver (round $1$): response\}\\[-3pt]
\dots\\[-0pt]
\{Reasoner (round $N$): thought \& question\}\\[0pt]
\{Perceiver (round $N$): response\}\par
\{Reasoner (round $N+1$): final answer\}\\[0pt]
\textless/think\textgreater\ \{Final answer from Reasoner\}
\end{tcolorbox}

\subsubsection{Self-play Prompting Mechanism}
\label{sec:self-play}

However, in practice, we found that the Reasoner tends to forgo detailed reasoning, often defaulting to sending the original query directly to the Perceiver, resulting in an overly ``jumpy'' reasoning process. The Perceiver, in turn, is prone to returning overly exhaustive visual information, much of it irrelevant to the specific question. This distracts the Reasoner, leading it to produce hasty responses. 

Therefore, we designed an elegant self-play prompting mechanism. Specifically, we prompt each agent that its collaborator is ``weak''. The Reasoner is told, ``The Perceiver you are collaborating with cannot understand complex questions.'' The Perceiver is told, ``The Reasoner you are collaborating with has weak reasoning abilities.'' This strategy forces the Reasoner to perform detailed problem decomposition and pose simple, incremental questions. Concurrently, it compels the Perceiver to provide answers that are accurate yet concise, without extraneous information. This self-play mechanism ensures their multi-turn dialogue converges into a detailed, progressively structured reasoning trace.

Ultimately, we synthesized a multi-modal RS-EoT dataset, termed \OurSFTData{}, (including RGB, infrared, and SAR imagery) for SFT cold-starting. Data statistics and training details are provided in \cref{sec:exp}.

\subsection{RL: Enhancing and Generalizing RS-EoT}
\label{sec:two_stage_rl}


To ensure the model's RS-EoT reasoning capability is sufficiently powerful, we build upon the SFT stage with a two-stage progressive reinforcement learning pipeline designed to first enhance and subsequently generalize this capability.

\subsubsection{Stage\#1: Fine-grained Grounding RL}
\label{sec:grounding_rl}

Starting with the SFT model, adhering to the principle that ``iron sharpens iron,'' we first perform reinforcement learning on fine-grained grounding tasks. These tasks are selected precisely because they most demand fine-grained visual evidence-seeking, thereby specifically strengthening the model's RS-EoT reasoning capability. 

Implementation-wise, we prompt the model to output the target object's location in the format ``[x1, y1, x2, y2]''. We then calculate the Intersection over Union (IoU) with the ground truth and use the IoU score directly as the reward. We also apply a strict format-based reward during this process to ensure output consistency.

\subsubsection{Stage\#2: General RS VQA RL}
\label{sec:vqa_rl}

To generalize the model's reasoning to broader RS scenarios, we further perform RL on general RS VQA datasets. However, existing RS VQA datasets predominantly contain \emph{Yes/No} or low-reasoning-intensity questions~\cite{lobry2021rsvqa,luo2024skysensegpt,li2024hrvqa}, which are highly susceptible to reward hacking.

To address this, we propose a multiple-choice RL data reconstruction strategy that converts existing simple RS VQA datasets into a format that stably supports RL training. Specifically, we observe that while current RS VQA datasets feature simple questions and answers, they often map a single image to multiple QA pairs (stemming from their construction process, often being programmatically converted from traditional classification or detection datasets). Leveraging this property, we convert them into multiple-choice questions. Without loss of generality, the reconstruction process for a single sample is as follows:

\begin{enumerate}[leftmargin=1.4em,label=\arabic*.,itemsep=2pt,topsep=2pt]
\item Collect a remote sensing image \(I\) and its associated QA set \(\{(Q_i, A_i)\}_{i=1}^{m}\) with \(10<m<15\).
\item Randomly invert the answers of \(n\) QA pairs to create \(n\) incorrect QA pairs, for example by changing ``Yes'' to ``No'' or adding/subtracting a random integer from a numeric answer, with \(1<n<m\).
\item Formulate a multiple-choice question (MCQ) sample using \textit{``Which of the following QA pairs match this remote sensing image?''} as the query, and use the \(m\) QA pairs as the options.
\end{enumerate}

The following is an example:

\begin{tcolorbox}[enhanced,
  colback=white,colframe=black!20,boxrule=0.4pt,arc=1pt,
  left=5pt,right=5pt,top=2pt,bottom=2pt,
  title={Multiple-Choice RL Data Reconstruction Example}, fonttitle=\bfseries,
  fontupper=\ttfamily\footnotesize]
\setlength{\parskip}{1pt}\setlength{\parindent}{0pt}
\textbf{Query:}\textless image\textgreater Which of the following QA pairs match this remote sensing image?\par
A. Is there a square in the image? yes\\[-0pt]
B. Are there cars in the image? no \\[-0pt]
\textbf{\textit{\# Inverted; cars are actually present}}\\[-0pt]
\dots\\[-0pt]
J. How many ships are in the image? 3\\[-0pt]
K. How many trees are in the image? 7 \\[-0pt]
\textbf{\textit{\# Inverted; there are actually 4 trees}}\par
\textbf{Ground truth:} A, ..., E, J
\end{tcolorbox}

\noindent\textbf{Reward design.} Given the large number of options, the model often cannot produce a perfectly correct answer on this reconstructed data. To avoid training collapse, we design a tailored graded reward function. Specifically, we assign an option-level, equal-weight, symmetric accuracy reward: \emph{selecting a correct option} and \emph{correctly rejecting an incorrect option} both receive a positive reward, whereas \emph{selecting an incorrect option} or \emph{failing to select a correct option} receive zero reward. Let $\mathbf{y}\in\{0,1\}^N$ denote ground-truth labels (1 = correct, 0 = incorrect) and $\hat{\mathbf{y}}\in\{0,1\}^N$ the model's binary choices (1 = chosen, 0 = not chosen). The per-question reward is
\begin{equation}
r_{\mathrm{qa}}
= 1 - \frac{1}{N}\sum_{i=1}^{N} \left|y_i-\hat{y}_i\right|.
  \label{eq:reward}
\end{equation}
This reward provides a stable, graded training signal by \emph{symmetrically} penalizing both misses and false selections and \emph{equally} weighting all options. This design effectively mitigates reward hacking and necessitates multi-round reasoning with evidence aggregation, forcing the model to verify each distinct option against the visual evidence to maximize its score.

\definecolor{oursC}{HTML}{DAE8FC}   
\definecolor{qwenC}{HTML}{ECF4FF}   
\definecolor{grpC}{HTML}{F5F5F5}    
\definecolor{newC}{HTML}{E6F4EA}    

\begin{table*}[]
\centering
\caption{Main comparison on remote sensing general VQA and fine-grained grounding. We compare \OurModelName{} with recent multimodal reasoning models (all based on Qwen2.5-VL-7B) as well as two remote-sensing-oriented reasoning models (Geo-R1 and VHM-RL). \OurModelName{} consistently achieves the best results across both VQA and grounding tasks, demonstrating the effectiveness of the RS-EoT paradigm in remote sensing scenarios. Abbreviations: VLAA-T = VLAA-Thinker, VL-R = VL-Rethinker, Vis-R1 = Vision-R1, MM-E = MM-Eureka, R1-OV = R1-OneVision.}
\vspace{-5pt}
\resizebox{\textwidth}{!}{%
\begin{tabular}{l c c c c c c c c c c c}
\toprule
\textbf{Benchmark} & \textbf{Metric} &
\cellcolor{oursC}\textbf{\OurModelName} &
\cellcolor{qwenC}\textbf{Qwen2.5VL}~\cite{bai2025qwen2} &
\textbf{WeThink}~\cite{yang2025wethink} &
\textbf{VLAA-T}~\cite{chen2025sft} &
\textbf{VL-R}~\cite{wang2025vl} &
\textbf{Vis-R1}~\cite{huang2025vision} &
\textbf{MM-E}~\cite{meng2025mm} &
\textbf{R1-OV}~\cite{yang2025r1} &
\cellcolor{newC}\textbf{Geo-R1}~\cite{geo-r1} &
\cellcolor{newC}\textbf{VHM-RL}~\cite{vhm-rl} \\
\midrule
\multicolumn{12}{c}{\cellcolor{grpC}\textit{Remote Sensing General QA}} \\
\midrule
\multirow{3}{*}{\textbf{RSFG-VQA}~\cite{luo2024skysensegpt}}
& Avg@5  & \cellcolor{oursC}\textbf{67.85} & \cellcolor{qwenC}62.45 & 55.04 & 55.72 & 58.80 & 39.95 & 60.98 & 47.70 & \cellcolor{newC}45.03 & \cellcolor{newC}53.04 \\
& Conv@5 & \cellcolor{oursC}\textbf{68.90} & \cellcolor{qwenC}62.46 & 41.07 & 37.87 & 57.88 & 41.94 & 60.82 & 48.00 & \cellcolor{newC}39.76 & \cellcolor{newC}51.06 \\
& Pass@5 & \cellcolor{oursC}\textbf{85.28} & \cellcolor{qwenC}77.16 & 61.43 & 77.88 & 76.37 & 57.81 & 71.22 & 77.66 & \cellcolor{newC}71.96 & \cellcolor{newC}61.56 \\
\midrule
\multirow{2}{*}{\textbf{RSFG-SC}~\cite{luo2024skysensegpt}}
& Scene@acc & \cellcolor{oursC}64.05 & \cellcolor{qwenC}57.42 & 60.12 & 56.50 & 58.82 & 21.63 & 56.98 & 47.04 & \cellcolor{newC}52.46 & \cellcolor{newC}\textbf{64.90} \\
& Object@F1 & \cellcolor{oursC}\textbf{56.52} & \cellcolor{qwenC}36.78 & 38.35 & 32.91 & 34.84 & 6.27 & 31.57 & 24.65 & \cellcolor{newC}20.82 & \cellcolor{newC}30.89 \\
\midrule
\multirow{3}{*}{\textbf{VRSBench}~\cite{li2024vrsbench}}
& Avg@5  & \cellcolor{oursC}\textbf{63.09} & \cellcolor{qwenC}62.45 & 62.17 & 60.54 & 55.04 & 37.72 & 55.25 & 44.59 & \cellcolor{newC}57.00 & \cellcolor{newC}57.01 \\
& Conv@5 & \cellcolor{oursC}\textbf{64.12} & \cellcolor{qwenC}62.58 & 62.36 & 61.16 & 55.49 & 37.39 & 55.37 & 46.36 &\cellcolor{newC}58.65 & \cellcolor{newC}56.76 \\
& Pass@5 & \cellcolor{oursC}\textbf{83.54} & \cellcolor{qwenC}75.62 & 73.84 & 77.57 & 71.27 & 48.91  & 66.85 & 73.95 & \cellcolor{newC}81.42 & \cellcolor{newC}70.03 \\
\midrule
\multirow{3}{*}{\textbf{RSVQA}~\cite{lobry2020rsvqa}}
& Avg@5  & \cellcolor{oursC}\textbf{75.16} & \cellcolor{qwenC}67.20 & 40.74 & 42.18 & 65.57 & 57.99 & 67.78 & 46.34 & \cellcolor{newC}34.50 & \cellcolor{newC}67.29 \\
& Conv@5 & \cellcolor{oursC}\textbf{78.29} & \cellcolor{qwenC}67.45 & 41.10 & 47.76 & 66.31 & 61.03 & 67.97 & 51.12 & \cellcolor{newC}45.78 & \cellcolor{newC}67.81 \\
& Pass@5 & \cellcolor{oursC}\textbf{92.51} & \cellcolor{qwenC}77.95 & 54.86 & 70.64 & 79.20 & 76.71 & 75.86 & 82.34 & \cellcolor{newC}65.43 & \cellcolor{newC}72.14 \\
\midrule
\multicolumn{12}{c}{\cellcolor{grpC}\textit{Fine-grained Grounding}} \\
\midrule
\multirow{3}{*}{\textbf{DIOR-RSVG}~\cite{zhan2023rsvg}}
& IoU@50 & \cellcolor{oursC}\textbf{47.00} & \cellcolor{qwenC}35.40 & 34.51 & 7.16 & 24.28 & 3.21 & 2.73 & 4.37 & \cellcolor{newC}17.67 & \cellcolor{newC}44.63 \\
& IoU@70 & \cellcolor{oursC}\textbf{33.32} & \cellcolor{qwenC}20.84 & 20.76 & 0.93 & 13.35 & 0.43 & 0.20 & 1.29 & \cellcolor{newC}7.21 & \cellcolor{newC}29.37 \\
& mIoU   & \cellcolor{oursC}\textbf{45.29} & \cellcolor{qwenC}35.64 & 33.96 & 12.84 & 25.48 & 8.91 & 6.87 & 7.82 & \cellcolor{newC}20.97 & \cellcolor{newC}40.32 \\
\midrule
\multirow{3}{*}{\textbf{VRSBench-Ref}~\cite{li2024vrsbench}}
& IoU@50 & \cellcolor{oursC}\textbf{54.71} & \cellcolor{qwenC}19.10 & 35.56 & 26.46 & 24.69 & 3.5 & 14.64 & 1.97 & \cellcolor{newC}17.18 & \cellcolor{newC}33.13 \\
& IoU@70 & \cellcolor{oursC}\textbf{32.40} & \cellcolor{qwenC}7.58 & 18.74 & 11.95 & 11.83 & 0.63 & 6.33 & 0.36 & \cellcolor{newC}13.77 & \cellcolor{newC}35.40 \\
& mIoU   & \cellcolor{oursC}\textbf{48.04} & \cellcolor{qwenC}21.99 & 34.07 & 27.55 & 25.29 & 8.62 & 15.69 & 5.48 & \cellcolor{newC}4.51 & \cellcolor{newC}18.74 \\
\bottomrule
\end{tabular}%
}
\vspace{-5pt}
\label{tab:main_res}
\end{table*}

\section{Experiments}
\label{sec:exp}

\noindent \textbf{Experimental Design.} Our experiments are designed to (1) evaluate the effectiveness of the RS-EoT paradigm and (2) provide in-depth analyses of its mechanisms. For evaluation, we conduct quantitative comparisons against SOTA models on VQA and grounding tasks (\cref{sec:exp_comp_mmrm}). We also compare our SocraticAgent data against traditional distillation methods (\cref{sec:exp_comp_data}). For analysis, we provide case studies illustrating \OurModelName{}'s iterative evidence-seeking (\cref{sec:exp_case}) and analyze attention dynamics to show its reasoning-perception cycles (\cref{sec:attention}). We also present the VQA-RL reward curve to validate our data strategy (\cref{sec:vqa_rl_curve}). Finally, we conduct a stage-wise ablation study to verify the contribution of each component (\cref{sec:training_stage}).

\noindent \textbf{Implementation details.} We set a 6-round dialogue limit for SocraticAgent and synthesize 4.3K SFT samples from the datasets listed in \cref{tab:data_src}. SFT training is conducted for 5 epochs with a learning rate of \(3\times10^{-5}\) using LLaMA-Factory~\cite{zheng2024llamafactory}. The two-stage RL procedure is implemented with EasyR1~\cite{zheng2025easyr1} and GRPO~\cite{shao2024deepseekmath}. In Stage 1, the model is trained on DIOR-RSVG~\cite{zhan2023rsvg} and VRSBench~\cite{li2024vrsbench}. In Stage 2, the model is trained on RS-VQA~\cite{lobry2021rsvqa} and FIT-RS~\cite{luo2024skysensegpt}. Both RL stages are trained for 2 epochs with a batch size of 512 and a learning rate of \(10^{-6}\).

\begin{table}[t]
\centering
\caption{Statistics of the data sources used for synthesizing our RS-EoT dataset.}
\vspace{-5pt}
\small
\begin{tabular}{lcc}
\toprule
\textbf{Data Source} & \textbf{Modality} & \textbf{Count} \\
\midrule
FIT-RS~\cite{luo2024skysensegpt} & RGB & 1.9K \\
VRSBench~\cite{li2024vrsbench} & RGB & 1.1K \\
DroneVehicle~\cite{sun2022drone} & RGB \& INF & 0.2K \\
SARLang-1M~\cite{wei2025sarlang} & SAR & 0.2K \\
EarthVQA~\cite{earthvqa} & RGB & 0.6K \\
RSVQA~\cite{lobry2020rsvqa} & RGB & 0.3K \\
\midrule
\textbf{Total} & \textbf{RGB/INF/SAR} & \textbf{4.3K} \\
\bottomrule
\end{tabular}
\vspace{-10pt}
\label{tab:data_src}
\end{table}

\noindent \textbf{Benchmark.} We evaluate on \textit{Remote Sensing General QA} (FiT-RSFG-VQA~\cite{luo2024skysensegpt}, FiT-RSFG-SC~\cite{luo2024skysensegpt}, VRSBench-VQA~\cite{li2024vrsbench}, RSVQA~\cite{lobry2020rsvqa}) and \textit{Fine-grained Grounding} (DIOR-RSVG~\cite{zhan2023rsvg}, VRSBench-Ref~\cite{li2024vrsbench}).

\noindent \textbf{Metrics.} For VQA, we report Avg@5, Conv@5, and Pass@5, which evaluate the average correctness, majority correctness, and any-correct correctness among 5 generated answers, respectively. We use accuracy for classification and F1 score for multi-object recognition. For grounding, we report mIoU and IoU@N (N=50, 70), the percentage of predictions with IoU \(>\) 0.5 or 0.7, respectively.

\noindent \textbf{Baseline.} We compare against SOTA reasoning models (\eg, WeThink~\cite{yang2025wethink}, Vision-R1~\cite{huang2025vision}, R1-OneVision~\cite{yang2025r1}) built on Qwen2.5-VL-7B~\cite{bai2025qwen2}, and RS-specific models (Geo-R1~\cite{geo-r1}, VHM-RL~\cite{vhm-rl}). For data comparison, we test distillation from Qwen3-VL-plus, Doubao-seed-1-6-vision, and GLM-4.5V. We exclude Gemini/GPT-5 models as their internal processes are inaccessible.

\begin{figure*}[]
  \centering
  \includegraphics[width=0.95\linewidth]{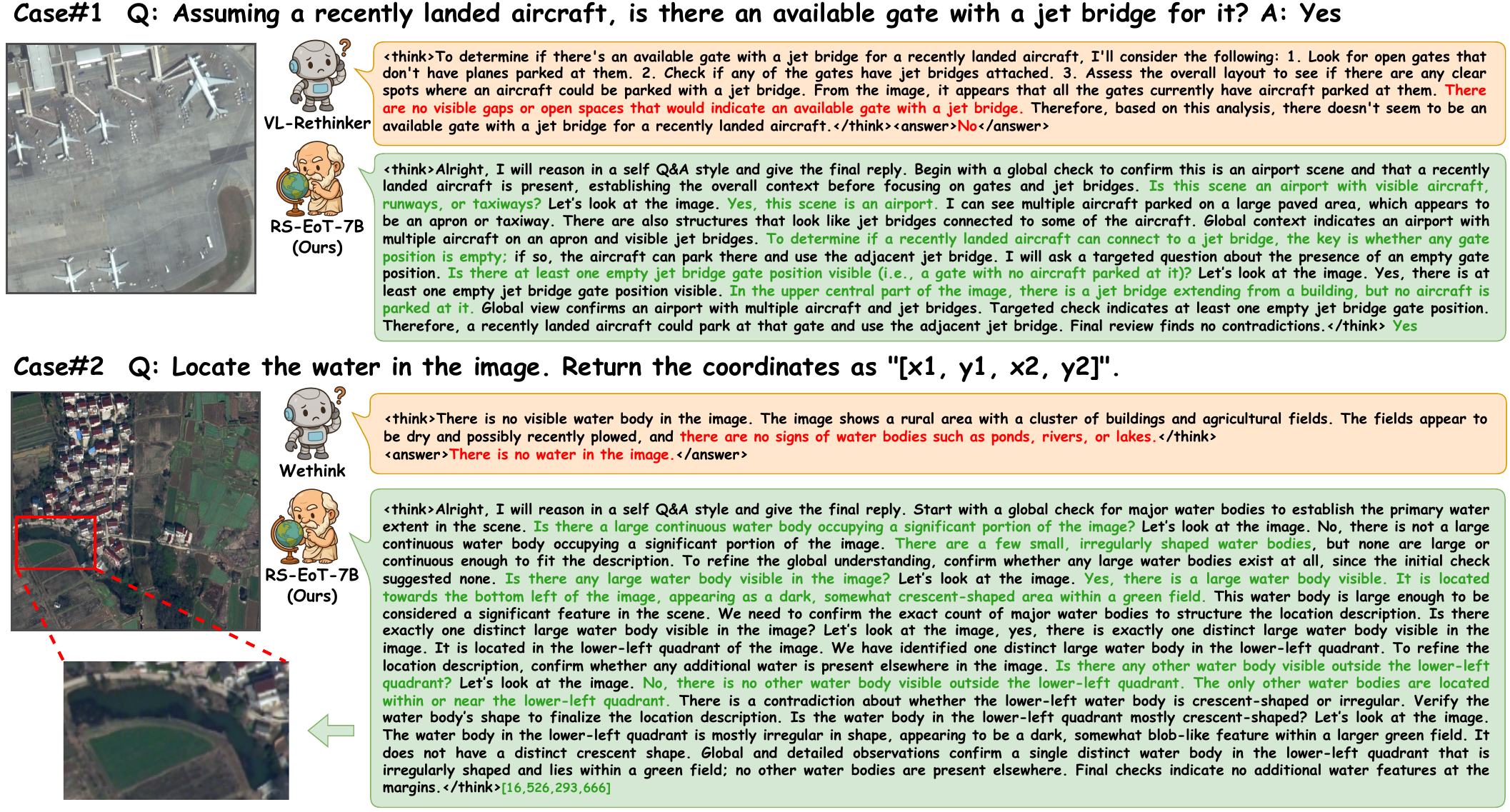}
  \vspace{-7pt}
  \caption{Case studies comparing \OurModelName{} with prior multimodal reasoning models on (top) Remote Sensing General QA and (bottom) Fine-grained Grounding. Unlike previous models, \OurModelName{} follows the RS-EoT paradigm: it iteratively self-questions, gathers additional visual evidence during reasoning, and uses that evidence to verify or adjust its conclusions.}
  \vspace{-10pt}
  \label{fig:case_study}
\end{figure*}

\definecolor{oursC}{HTML}{DAE8FC}
\begin{table}[]
\caption{SFT performance comparison of reasoning trace synthesis strategies on FIT-RSFG-VQA. Traces generated by SocraticAgent yield stronger reasoning performance than direct distillation from frontier models (Qwen3VL, Doubao, GLM-4.5V).}
\vspace{-5pt}
\centering
\small
\resizebox{0.95\linewidth}{!}{%
\begin{tabular}{l ccc}
\toprule
\textbf{Method} & \textbf{Avg@5} & \textbf{Conv@5} & \textbf{Pass@5} \\
\midrule
\cellcolor{oursC}\textbf{SocraticAgent (ours)} & \cellcolor{oursC}64.82 & \cellcolor{oursC}67.35 & \cellcolor{oursC}85.89 \\
Qwen3-VL-plus~\cite{team2025qwen3}   & 63.78 & 65.15 & 76.87 \\
Doubao-seed-1-6-vision~\cite{doubao-seed-1.6}    & 51.80 & 17.17 & 66.81 \\
GLM-4.5V~\cite{vteam2025glm45vglm41vthinkingversatilemultimodal}  & 66.25 & 66.58 & 77.99 \\
\bottomrule
\end{tabular}%
}
\vspace{-10pt}
\label{tab:distill_compare}
\end{table}

\subsection{Comparison with Prior Reasoning Models}
\label{sec:exp_comp_mmrm}

\cref{tab:main_res} compares \OurModelName{} with recent multimodal reasoning models and two RS specific reasoning models across both remote sensing VQA and fine-grained grounding benchmarks. Except for Geo-R1 and VHM-RL, all competing models are built on Qwen2.5-VL-7B, ensuring a fair comparison focused on the effectiveness of the reasoning paradigm rather than the capacity of the base model.

On the remote sensing VQA benchmarks, \OurModelName{} consistently achieves the best performance across all three evaluation metrics (Avg@5, Conv@5, Pass@5), indicating stronger reasoning ability and more stable answer quality. In contrast, existing multimodal reasoning models exhibit noticeable instability: their performance fluctuates significantly relative to the base model and fails to deliver consistent gains on remote sensing tasks, suggesting that their generated reasoning traces primarily reflect \emph{pseudo reasoning} rather than genuine evidence-guided reasoning.

For fine-grained grounding tasks, \OurModelName{} also surpasses all baselines by a large margin in IoU@50, IoU@70, and mIoU. These improvements demonstrate that the iterative evidence-seeking mechanism effectively guides the model's attention to and improves its ability to localize visual cues.

\subsection{Comparison with Frontier Model Distillation}
\label{sec:exp_comp_data}

\begin{figure*}[]
  \centering
  \includegraphics[width=\linewidth]{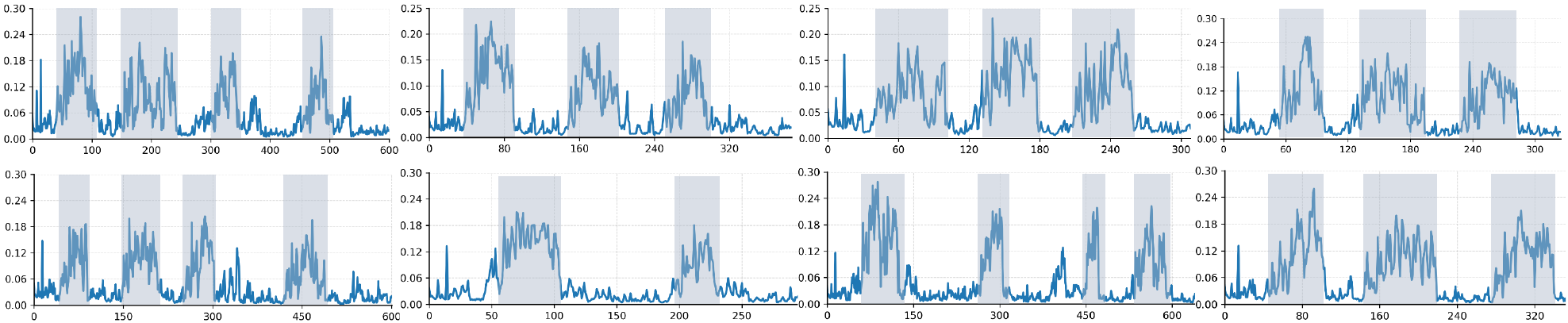}
  \vspace{-17pt}
  \caption{Token-wise attention visualization on eight randomly sampled cases. The y-axis represents the proportion of attention allocated to image tokens, and the x-axis represents the token index during the decoding step. Clear periodic patterns emerge: attention peaks on visual tokens (evidence-seeking phases) and then drops during language-based reasoning (reasoning phases). This alternating cycle reflects the iterative reasoning mechanism instilled by the RS-EoT paradigm.}
  \vspace{-5pt}
  \label{fig:attn_analysis}
\end{figure*}

We compare our SocraticAgent against the common practice of directly distilling reasoning traces from frontier models. We sampled 2K queries from the FIT-RSFG-VQA training set and generated reasoning traces in two ways: (1) using our SocraticAgent, and (2) via direct distillation from Qwen3-VL-plus, Doubao-seed-1-6-vision, and GLM-4.5V. We then fine-tuned models on each dataset and evaluated them on FIT-RSFG-VQA, with results shown in \cref{tab:distill_compare}.

The results show that models trained on SocraticAgent-generated traces consistently outperform those trained on directly distilled traces. This suggests that frontier models often do not expose reasoning processes aligned with remote sensing cognition. In contrast, SocraticAgent synthesizes structured, iterative, and evidence-seeking trajectories, leading to more reliable and transferable reasoning behavior in remote sensing.

\subsection{Case Study}
\label{sec:exp_case}

We present two qualitative examples in \cref{fig:case_study} to illustrate the RS-EoT paradigm. In the VQA example, the baseline performs only a single, coarse scene interpretation and fails to find an available jet bridge, missing contradictory visual evidence. In contrast, \OurModelName{} performs iterative verification: it first establishes the airport context, then explicitly searches for an \textit{unoccupied} gate, and finally identifies an available one. In the grounding example, the baseline fails to detect the small, irregular water body. \OurModelName{}, however, uses iterative inspection guided by self-questioning: it first performs a global scan, then identifies a candidate region (lower-left), and re-examines the image to rule out other water bodies. This sequence of targeted visual checks allows the model to refine both what is present and where it is located, ultimately producing accurate coordinates supported by its grounded reasoning.

\begin{figure}[]
  \centering
  \includegraphics[width=0.90\linewidth]{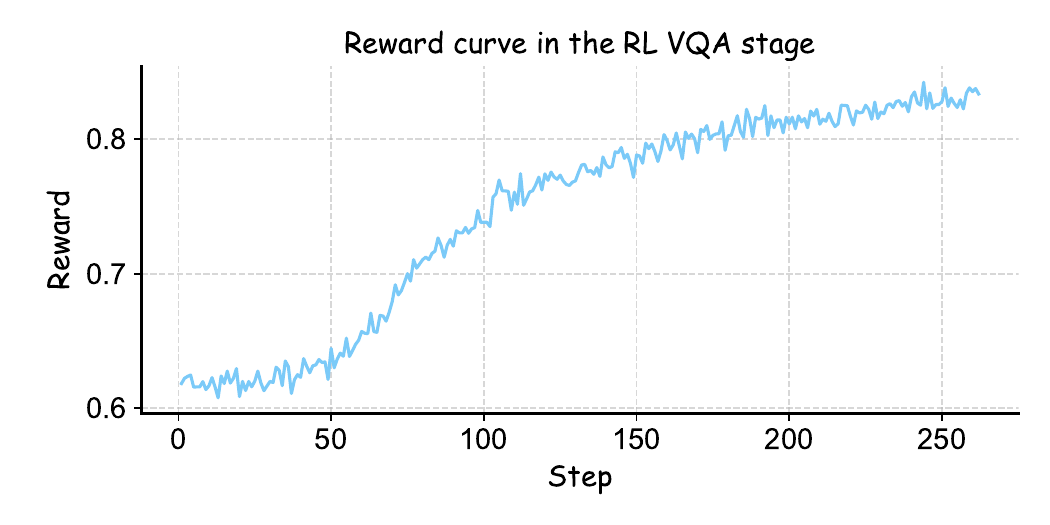}
  \vspace{-10pt}
  \caption{The reward curve for the VQA RL stage. The stable upward trend validates that our multiple-choice data reconstruction strategy provides an effective learning signal and successfully mitigates reward hacking.}
  \vspace{-5pt}
  \label{fig:RLVQARewards}
\end{figure}

\subsection{Attention Dynamics During EoT Reasoning}
\label{sec:attention}

To understand how RS-EoT guides the model to alternate between linguistic reasoning and visual evidence seeking, we visualize the token-wise attention distribution during decoding. For each generated token, we compute the proportion of its attention allocated to image tokens (y-axis) plotted against its index in the generated sequence (x-axis).

\cref{fig:attn_analysis} presents visualizations on eight randomly sampled test cases, all revealing a clear periodic pattern. Attention periodically peaks on visual tokens, which we interpret as \emph{evidence-seeking phases} where the model actively retrieves new visual cues from the image. These peaks are followed by troughs where attention shifts back to language tokens, corresponding to \emph{reasoning phases} for linguistic deduction. This alternating behavior aligns precisely with our RS-EoT paradigm, demonstrating that the model dynamically coordinates between image inspection and high-level inference, rather than merely generating text sequentially. These findings confirm that RS-EoT enables a structured, iterative reasoning process where the model repeatedly examines the image, updates its hypothesis, and advances its reasoning.

\subsection{VQA-RL Training Stability Analysis}
\label{sec:vqa_rl_curve}

\cref{fig:RLVQARewards} illustrates the reward curve during the VQA RL stage (Stage\#2). We observe a smooth and consistent upward trajectory, with the reward steadily increasing from an initial value of approximately 0.62 and converging around 0.84. This stable progression provides strong evidence for the effectiveness of our multiple-choice data reconstruction strategy. It demonstrates that our method, combined with the tailored graded reward function, successfully mitigates reward hacking and provides a stable, effective learning signal, enabling the model to learn from simple RS VQA datasets without training collapse.

\subsection{Stage-wise Training Analysis}
\label{sec:training_stage}

\definecolor{SFTC}{HTML}{E7F1FF}   
\definecolor{RLC}{HTML}{FFF2E6}    

\begin{table}[t]
\centering
\small
\caption{Ablation study of training stages on VQA and Grounding and the results validate the complementary role of each stage.}
\vspace{-5pt}
\resizebox{\linewidth}{!}{%
\begin{tabular}{l ccc ccc}
\toprule
\multirow{2}{*}{\textbf{Stage}} &
\multicolumn{3}{c}{\textbf{VQA}} &
\multicolumn{3}{c}{\textbf{Grounding}} \\
\cmidrule(lr){2-4}\cmidrule(lr){5-7}
& Avg@5 & Conv@5 & Pass@5 & IoU@50 & IoU@70 & mIoU \\
\midrule
Base     & 67.20 & 67.45 & 77.95 & 35.40 & 20.84 & 35.64 \\
\rowcolor{SFTC}
+SFT      & 70.73 & 74.05 & 91.96 &  8.81 &  2.19 & 13.95 \\
\rowcolor{RLC}
+RL-IoU   & 69.51 & 72.01 & 90.63 & 48.41 & 33.51 & 45.57 \\
\rowcolor{RLC}
+RL-VQA   & 75.16 & 78.29 & 92.51 & 48.23 & 33.27 & 45.52 \\
\bottomrule
\end{tabular}%
}
\vspace{-10pt}
\label{tab:stage_compare}
\end{table}

We conduct a stage-wise ablation to analyze the contribution of each training step, as shown in \cref{tab:stage_compare}. Starting from the Base model, SFT with RS-EoT trajectories substantially improves VQA accuracy but causes a significant drop in grounding ability. The subsequent RL-IoU stage effectively restores and dramatically enhances grounding performance, confirming that explicit spatial feedback is crucial for localization. Finally, the RL-VQA stage further improves VQA reasoning while successfully maintaining the strong grounding performance. These results highlight the complementary role of each stage: SFT injects the reasoning pattern, RL-IoU enhances the visual evidence-seeking capability, and RL-VQA allows the model to be applied to broader remote sensing scenarios.
\section{Conclusion}
\label{sec:conclusion}


In this work, we introduce RS-EoT, an iterative evidence-seeking paradigm for remote sensing understanding. RS-EoT structures reasoning as an alternating loop of linguistic deduction and targeted visual evidence seeking. We instill this paradigm using an SFT and progressive RL (Grounding+VQA) pipeline. Experiments show our approach surpasses SOTA baselines in reasoning robustness and grounding accuracy. Analyses confirm the model internalizes this iterative pattern, validating structured reasoning for trustworthy geospatial AI.
{
    \small
    \bibliographystyle{ieeenat_fullname}
    \bibliography{main}
}
\clearpage
\setcounter{page}{1}
\maketitlesupplementary


\section{System Prompts Details}
\label{app:prompts}

In this section, we provide the exact system prompts used in our SocraticAgent framework to synthesize the RS-EoT-4K dataset. As described in the main paper, SocraticAgent operates as a self-play multi-agent system consisting of three distinct roles: the \textbf{Reasoner}, the \textbf{Perceiver}, and the \textbf{Verifier}.

\begin{itemize}

\item \textbf{The Reasoner} (\cref{fig:reasoner_prompt}) serves as the cognitive core. It is a text-only agent designed to simulate the human ``coarse-to-fine'' visual interpretation process. By following a \textit{Plan–Integrate–Decide} paradigm, it decomposes complex queries into atomic visual questions, iteratively gathering evidence before concluding.

\item \textbf{The Perceiver} (\cref{fig:perceiver_prompt}) acts as the visual interface. To enforce explicit and detailed reasoning from the Reasoner, we employ a self-play prompting mechanism where the Perceiver is instructed to collaborate with a ``weak-reasoning'' partner. Its role is to provide accurate, descriptive visual observations in a ``natural inner monologue'' style without performing high-level logical reasoning.

\item \textbf{The Verifier} (\cref{fig:verifier_prompt}) functions as the quality control gatekeeper. It compares the final answer generated by the Reasoner against the ground truth to filter out incorrect or hallucinated reasoning traces during the data synthesis phase.

\end{itemize}

The complete prompts for these agents are presented in \cref{fig:reasoner_prompt}, \cref{fig:perceiver_prompt}, and \cref{fig:verifier_prompt}, respectively.

\begin{figure*}[]
    \centering
    \begin{promptbox}{System Prompt for the Reasoner}
    
    You are a reasoning model that follows a \textbf{Plan–Integrate–Decide} paradigm, collaborating with a \textbf{weak-perception visual model} to complete \textbf{general remote sensing tasks} (such as classification/attribute recognition, localization/counting, relation/change detection, and VQA). 
    The perception model can only answer \textbf{very simple, atomic visual facts} and cannot perform reasoning. 
    Therefore, you must decompose the perception process into a \textbf{coarse-to-fine sequence of steps}, simulating how humans visually interpret remote sensing imagery.
    
    \vspace{0.8em}
    \textbf{[Coarse-to-Fine Perception Chain]}
    \begin{enumerate}[leftmargin=*, label=\arabic*)]
        \item \textbf{Global Observation Stage (Overall Understanding):}
        \begin{itemize}[leftmargin=3mm, label=-]
            \item Begin with a \textbf{broad, holistic examination} of the entire image, forming an initial impression of its overall layout — main land-cover types, spatial organization, scene functionality, distribution of major objects, and possible visual interferences (e.g., shadows, fog, noise, or occlusion).
            \item While questions at this stage should remain \textbf{broad, general, and high-level}, they must be \textbf{context-aware} — i.e., lightly tailored to the task/query so they inform later reasoning for this specific problem.
        \end{itemize}
        
        \item \textbf{Focused and Detailed Observation Stage (Targeted Analysis):}
        \begin{itemize}[leftmargin=3mm, label=-]
            \item After forming a general understanding of the scene, use the task objective (query) and global observations to \textbf{focus attention} on potentially relevant local regions or objects.
            \item Naturally shift attention from overall impressions to specific, task-relevant areas, similar to how humans visually focus.
            \item Ask more detailed and targeted questions, typically focusing on:
            \begin{itemize}[leftmargin=3mm, label=$\bullet$]
                \item Local details (shape, texture, boundaries, orientation, color features, etc.);
                \item Relationships and differences (changes, similarities, transitions between regions, etc.);
                \item Task-critical elements (e.g., presence, quantity, or arrangement of specific targets).
            \end{itemize}
        \end{itemize}

        \item \textbf{Integration and Verification Stage:}
        \begin{itemize}[leftmargin=3mm, label=-]
            \item Integrate the facts collected from the global and detailed observation stages into a consistent intermediate conclusion.
            \item If contradictions or uncertainties remain, ask verification questions.
            \item Ensure that the reasoning covers all key regions and that the logic is consistent.
        \end{itemize}

        \item \textbf{Final Review and Confirmation Stage:}
        \begin{itemize}[leftmargin=3mm, label=-]
            \item Before giving the final answer, perform a quick overall review of the image to confirm that no small anomalies, marginal areas, or potential clues have been overlooked.
            \item Check whether the final answer meets the query's requirements regarding format, length, and structure.
            \item The final answer must only output the direct answer to the query itself, such as ``Yes/No'', a specific number, or a concise conclusion. Do not include any explanations, reasoning, or additional commentary.
            \item If necessary, ask one final targeted question for confirmation.
        \end{itemize}
    \end{enumerate}

    \vspace{0.8em}
    \textbf{[Questioning and Iteration Constraints]}
    \begin{itemize}[leftmargin=*, label=-]
        \item \textbf{Never} forward the user's original query directly to the perception model; each question must concern \textbf{only one atomic visual fact}.
        \item Each new question should provide \textbf{maximum information gain} and \textbf{must not repeat} previous questions (avoid paraphrasing).
        \item You have \texttt{\{MAX\_LOOP-1\}} questioning rounds available: the early rounds focus on global perception, the middle rounds gather key evidence, and the final rounds perform verification questioning.
    \end{itemize}

    \vspace{0.8em}
    \textbf{[Output Format (Strict Requirements)]}
    \begin{itemize}[leftmargin=*, label=-]
        \item \textbf{If further questioning is needed:} \\
        Start with \texttt{<thinking>...</thinking>} (briefly explain the reasoning and purpose of the next question), then output \textbf{only one} \texttt{<question>...</question>}.
        \item \textbf{If ready to give the final answer:} \\
        Start with \texttt{<thinking>...</thinking>} (summarize key evidence and note that final checks have been completed), then output \texttt{\{FINAL\_PREFIX\_EN\} ...}.
        \item Each round must \textbf{begin} with \texttt{<thinking>...</thinking>} and be followed by \textbf{exactly one} of the two options: \texttt{<question>...</question>} or \texttt{\{FINAL\_PREFIX\_EN\} ...}. No other content is allowed.
        \item Inside \texttt{<thinking>}, do \textbf{not} mention external entities such as ``the perception model,'' ``user,'' or ``conversation.''
        \item Use \textbf{English} for internal reasoning and questioning, but ensure that the \textbf{final answer matches the input query's language}.
    \end{itemize}
    
    \end{promptbox}
    \vspace{-0.3cm}
    \caption{The system prompt for the Reasoner in SocraticAgent.}
    \label{fig:reasoner_prompt}
\end{figure*}

\begin{figure*}[]
    \centering
    \begin{promptbox}{System Prompt for the Perceiver}
    
    You are an image interpretation expert collaborating with a \textbf{reasoning model that has very weak logical ability}. Together, through multi-turn dialogue, you will complete \textbf{general remote sensing tasks} (classification/attribute, localization/counting, relation/change analysis, VQA, etc.).
    
    The reasoning model can \textbf{only understand the textual descriptions} of your perception results — it \textbf{cannot see the image directly}. Therefore, you must respond to each of its questions about the image \textbf{accurately and completely}, without adding any information that is irrelevant to the question.
    
    Your tone should resemble a \textbf{natural inner monologue} of a person carefully observing an image. Always begin your response with: ``Let's look at the image,'' and then continue with your detailed observation.
    
    \end{promptbox}
    \vspace{-0.3cm}
    \caption{The system prompt for the Perceiver in SocraticAgent.}
    \label{fig:perceiver_prompt}
\end{figure*}

\begin{figure}[]
    \centering
    \begin{promptbox}{System Prompt for the Verifier}
    
    You are a \textbf{strict answer evaluator}. Given a Query, Answer, and GT, output only:
    \begin{enumerate}[leftmargin=*, label=\arabic*)]
        \item \texttt{"ACCEPT"}
        \item \texttt{"REJECT: <brief reason>"}
    \end{enumerate}
    
    \end{promptbox}
    \vspace{-0.3cm}
    \caption{The system prompt for the Verifier in SocraticAgent.}
    \label{fig:verifier_prompt}
\end{figure}

\section{SFT Training Settings}
\label{app:sft_settings}

We perform SFT on the base model Qwen2.5-VL-7B-Instruct using the RS-EoT-4K dataset. The training is implemented based on the LLaMA-Factory framework. We train the model for 5 epochs with a learning rate of $3 \times 10^{-5}$, using the AdamW optimizer and a cosine learning rate scheduler. The global batch size is set to 64, and the maximum sequence length is limited to 4096 tokens to accommodate the detailed reasoning traces. The entire training process was executed on 4 A100 GPUs and required approximately 40 minutes to complete.

A empirical finding in our experiments concerns the use of system prompts. We observed that including explicit reasoning triggers in the system prompt (e.g., instructions like ``Please reason step-by-step'') during training caused the model to develop a strong dependency on these specific triggers. Consequently, the model often failed to initiate reasoning or exhibited abnormal behaviors when such prompts were absent during inference. To address this and ensure robust, spontaneous reasoning capabilities, we exclude all reasoning-specific instructions from the system prompt during SFT and RL. Instead, we hard-code the start-of-thought token \texttt{<think>} directly into the assistant's response field within the chat template. This design forces the model to automatically enter the reasoning mode immediately upon generation, ensuring stable output trajectories without relying on specific user or system prompts.

\section{RL Training Settings}
\label{app:rl_settings}

All reinforcement learning experiments are conducted using the EasyR1 framework, which provides a production-ready implementation of GRPO with KL regularization.
We fix the KL coefficient to $\beta = 1.0\times10^{-2}$.
For each input, the model generates 4 rollout samples using sampling temperature~1.0,
with a maximum response length of 4096 tokens.

Policy optimization is performed with the AdamW optimizer,
using a learning rate of $1.0\times10^{-6}$, weight decay of $1.0\times10^{-2}$, and
a linear warm-up over the first 3
We apply gradient clipping with a threshold of~1.0.
All models are trained in \texttt{bf16} precision under full-sharded data parallelism
(FSDP) with gradient checkpointing enabled.

For both RL-Grounding and RL-VQA, we use a global batch size of 512
(covering rollout, training, and validation batches) and train for 240 steps.
Experiments are run on a single node equipped with 8 A100 GPUs.
The RL-VQA stage requires approximately 2.3 days to complete, while
the RL-Grounding stage takes about 2.5 days.

\section{RL Reward Function}
\label{app:reward_function}

\subsection{Grounding Reward}
\label{app:reward_grounding}

For the grounding task, the model is required to output a bounding box in the
form \texttt{[x1, y1, x2, y2]} after a complete \texttt{<think></think>} block.
Our reward contains two components: an IoU-based accuracy term and a lightweight
format term.

\paragraph{Format reward.}
For the grounding task, we apply a lightweight \emph{format reward} to
encourage the model to produce both a well-formed reasoning block and a
structured bounding-box output.  
We evaluate two binary indicators:  
(i) whether the response contains a complete, matched
\texttt{<think></think>} block, and  
(ii) whether a valid four-number bounding-box list of the form
\texttt{[x1, y1, x2, y2]} appears after the closing \texttt{</think>} tag.  
We denote these indicators as $s_{\text{think}}$ and $s_{\text{bbox}}$,
respectively.  
The format reward is then defined as
\[
  r_{\text{fmt}} = \frac{s_{\text{think}} + s_{\text{bbox}}}{2},
\]

\paragraph{IoU accuracy.}
We first normalize the response and verify that a well-formed reasoning block
(i.e., a matched pair \texttt{<think>...</think>}) exists.
Only the text appearing \emph{after} the closing tag \texttt{</think>} is used
for prediction parsing.  
From this tail segment, we extract the first valid four-number list
\texttt{[x1, y1, x2, y2]}; if no valid list is found, the prediction is treated
as invalid.  
The accuracy score is then computed using the IoU between the predicted and
ground-truth boxes directly in the original pixel coordinate system:
\[
  r_{\text{acc}} =
  \begin{cases}
    \mathrm{IoU}(\text{bbox}_{\text{pred}}, \text{bbox}_{\text{gt}}), 
      & \text{if a valid box is parsed}, \\[2pt]
    0, & \text{otherwise}.
  \end{cases}
\]

\paragraph{Overall reward.}
The final grounding reward combines IoU accuracy and the format term:
\[
  r_{\text{overall}} = (1 - \lambda)\, r_{\text{acc}} 
  + \lambda\, r_{\text{fmt}},
\]
where we use $\lambda = 0.1$ in all experiments.
Each sample is scored independently, and we log
$r_{\text{overall}}$, $\mathrm{IoU}$, $r_{\text{acc}}$, and $r_{\text{fmt}}$
during training.

\subsection{Multiple-Choice VQA Reward}
\label{app:reward_vqa}

For the multiple-choice VQA task, the model predicts a subset of option letters
(e.g., \texttt{A}, \texttt{C}, \texttt{D}) after completing the
\texttt{<think></think>} reasoning block. 
Only the text following the last \texttt{</think>} tag is considered for answer
extraction. 
We first identify the last line in this tail that contains independent
letter tokens; if none are found, we treat the prediction as invalid.

Given the ground-truth correct set $G \subseteq \mathcal{A}$ and the model's
predicted set $P \subseteq \mathcal{A}$, where $\mathcal{A}$ is the set of all
allowed options, we compute a symmetric, option-level accuracy score.
Each option contributes equally: selecting a correct option and correctly
rejecting an incorrect option both receive positive credit, while selecting an
incorrect option or failing to select a correct one is penalized.  
The accuracy reward is therefore:

\[
r_{\text{acc}}
= 1 - \frac{1}{|\mathcal{A}|}
  \sum_{a \in \mathcal{A}}
  \big| \mathbf{1}\{a \in G\} - \mathbf{1}\{a \in P\} \big|.
\]

If the model outputs any letter not contained in $\mathcal{A}$, we apply a
hard-zero accuracy (i.e., $r_{\text{acc}} = 0$), matching our implementation.

\paragraph{Format reward.}
The VQA setting uses the same lightweight format reward as the grounding task.
Because empty outputs are allowed and interpreted as selecting no options, the
format reward does not require the predicted letters to appear after the
\texttt{</think>} tag. 
Instead, we simply check whether the model produces a complete and matched
\texttt{<think></think>} reasoning block. 
If the block is present, the model earns a format score of 1; otherwise, the
score is 0.

\paragraph{Overall reward.}
The final VQA reward is a convex combination of accuracy and format:
\[
r_{\text{overall}}
= (1 - \lambda)\, r_{\text{acc}}
+ \lambda\, r_{\text{fmt}},
\]
with $\lambda = 0.1$ throughout all experiments.

This reward design provides a graded, stable learning signal across all options
without relying on task-specific heuristics, while the matched
\texttt{<think></think>} constraint ensures structured reasoning outputs.

\section{RL Training Dynamics Curves}
\label{app:rl_dynamics}

\begin{figure}[t]
  \centering
    \includegraphics[width=\linewidth]{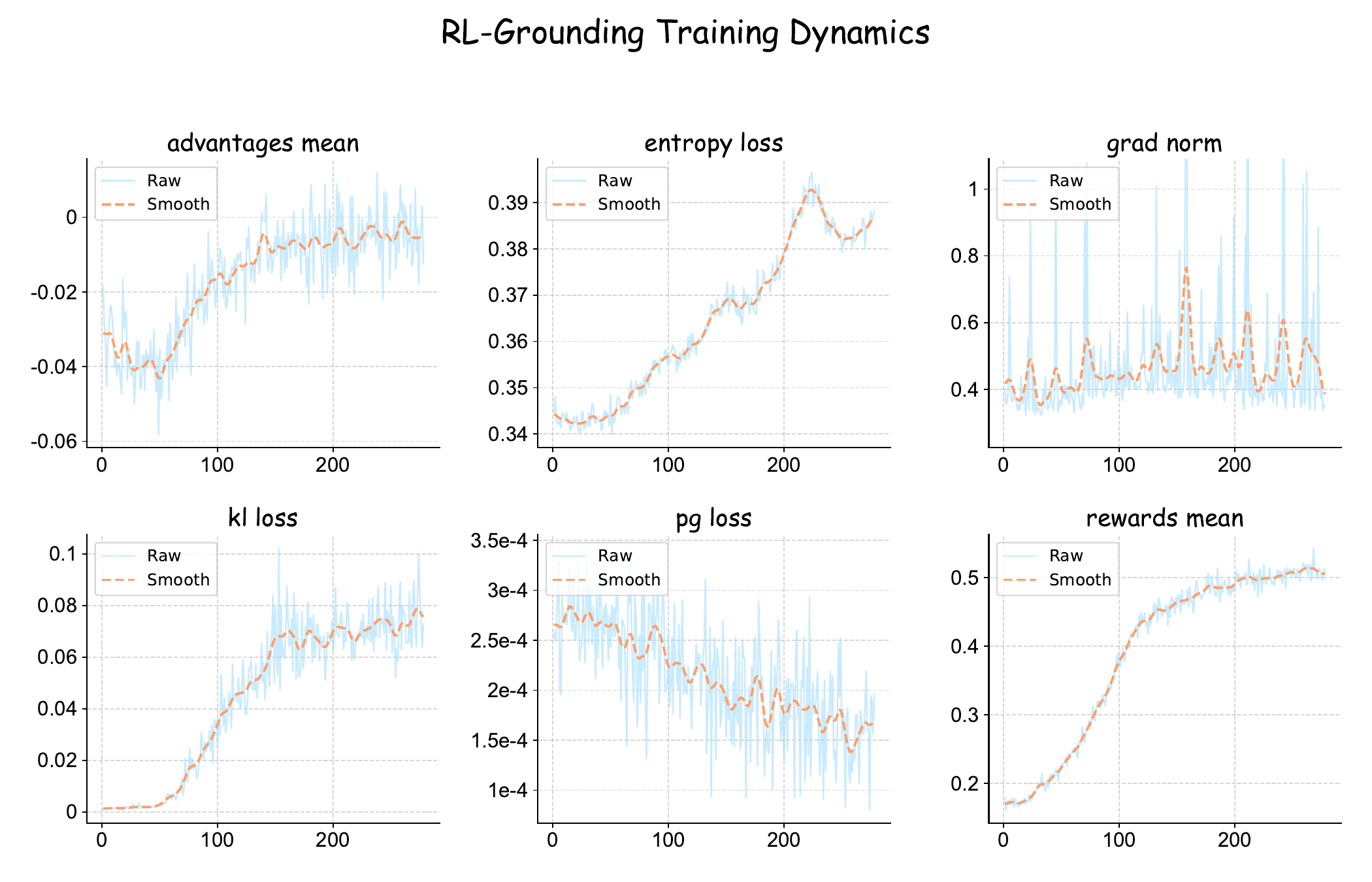}\\[1.5em]
    \includegraphics[width=\linewidth]{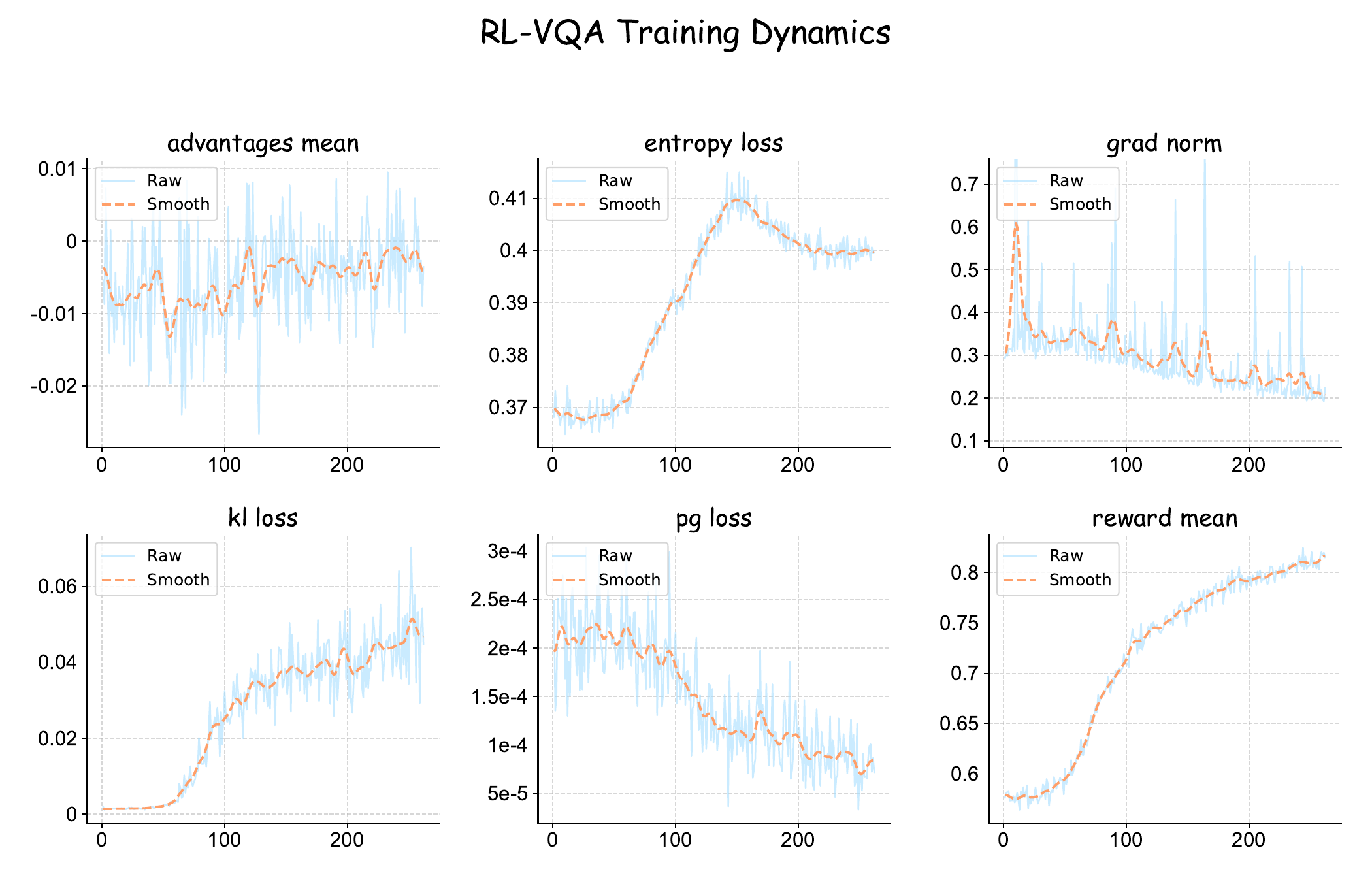}
    \caption{
        RL training dynamics for our two-stage pipeline.
        Top: RL-Grounding stage; bottom: RL-VQA stage.
        Each panel reports the evolution of a different statistic (
        advantages mean, entropy loss, gradient norm, KL loss, policy gradient
        loss, and rewards mean), with light curves showing raw values and
        dashed curves showing Gaussian-smoothed trends over training steps.
    }
    \label{fig:rl_training_dynamics}
\end{figure}

Figure~\ref{fig:rl_training_dynamics} visualizes the evolution of key
optimization statistics during the two RL stages in our pipeline: RL-Grounding and RL-VQA.  
The top block corresponds to the RL Grounding stage and the bottom block to the RL-VQA stage; in both cases we plot the same set of metrics, including mean advantage, actor gradient norm, entropy loss, KL loss, policy gradient loss, and average reward. 
For each quantity, we plot both the raw measurements and their
Gaussian-smoothed trends, providing a compact view of how the GRPO
optimization behaves over the training steps in both stages.

\section{Difference Between Multiple-Choice VQA and Standard VQA}
\label{app:vqa_ablation}

To assess the effectiveness of our proposed multiple-choice reformulation of VQA,
we additionally perform an ablation study using the original dataset and model
settings, but applying reinforcement learning directly on the \emph{standard}
free-form VQA answers.  
This experiment allows us to isolate and compare the impact of our
multiple-choice VQA design against the conventional VQA supervision.

\begin{figure}[]
    \centering
    \includegraphics[width=0.95\linewidth]{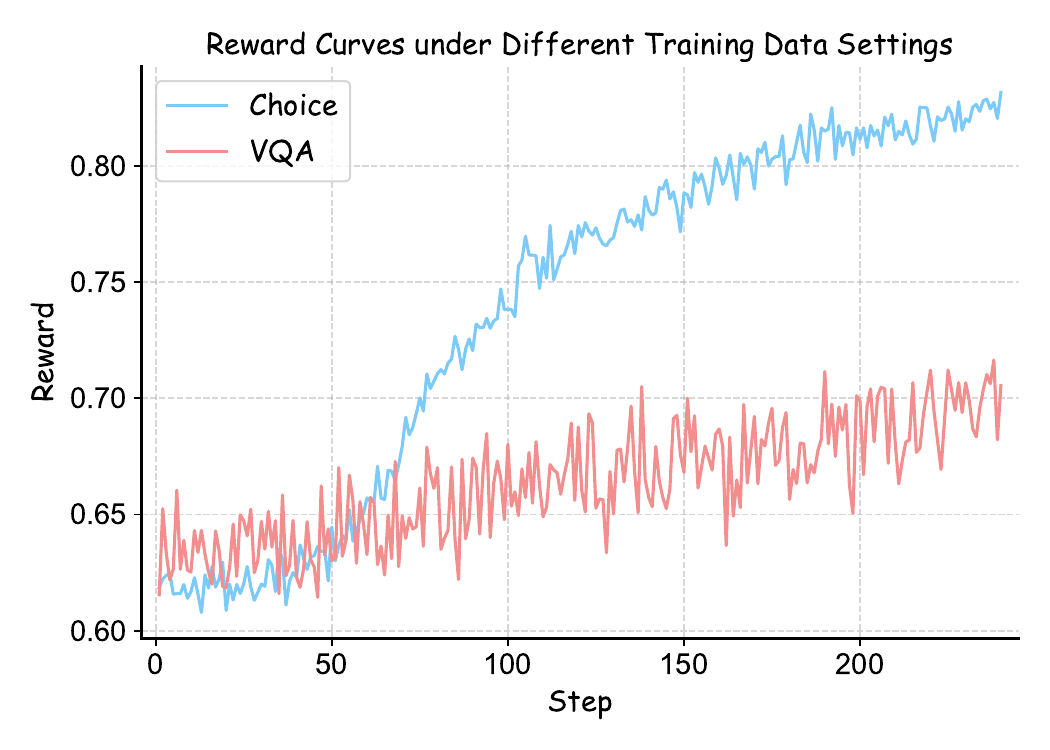}
    \caption{
        Ablation comparing reinforcement learning on the original
        free-form VQA answers (\textcolor{red}{red}) versus our
        reconstructed multiple-choice VQA reconstruction (\textcolor{blue}{blue}).
        Standard VQA produces unstable and oscillatory training dynamics,
        while the multiple-choice approach yields smooth and efficient reward
        improvement.
    }
    \label{fig:vqa_ablation_curve}
\end{figure}

We first evaluate the final model performance on the RSVQA. As presented in Table~\ref{tab:vqa_formulation_ablation}, our multiple-choice reconstruction strategy achieves consistently superior results across all metrics compared to the standard free-form supervision. 
To understand the underlying cause of this performance gap, we visualize the training stability in Figure~\ref{fig:vqa_ablation_curve}. When trained with standard
free-form VQA answers, the model exhibits slow reward improvement and frequent
large-magnitude oscillations during GRPO optimization.  
This indicates that the original VQA supervision produces an unstable and
inefficient reward signal for RL, making it difficult for the model to learn
consistent behaviors. In contrast, our multiple-choice VQA reformulation leads to a much smoother and
steadily increasing reward curve under the \emph{same} RL setup.  
This demonstrates that transforming VQA into a structured multi-option
prediction task greatly stabilizes the reward landscape and improves training
efficiency.

Overall, the ablation confirms that the multiple-choice VQA formulation provides a
more reliable and effective RL objective compared to standard VQA, and is
essential for achieving stable reinforcement learning in our setting.

\begin{table}[]
    \centering
    \caption{Ablation study on VQA data reconstruction strategies on the RSVQA. We compare our proposed Multiple-Choice VQA reconstruction against the Standard VQA (Free-form) supervision. Our method achieves consistently superior performance across all metrics.}
    \label{tab:vqa_formulation_ablation}
    \resizebox{\linewidth}{!}{
        \begin{tabular}{l ccc}
            \toprule
            \textbf{Method} & \textbf{Avg@5} & \textbf{Conv@5} & \textbf{Pass@5} \\
            \midrule
            Standard VQA (Free-form) & 74.73 & 76.86 & 90.91 \\
            \textbf{Ours (Multiple-Choice)} & \textbf{75.16} & \textbf{78.29} & \textbf{92.51} \\
            \bottomrule
        \end{tabular}
    }
\end{table}

\section{Case Study}
\label{app:case_supp}

\begin{figure*}[t]
  \centering
  \includegraphics[width=0.95\linewidth]{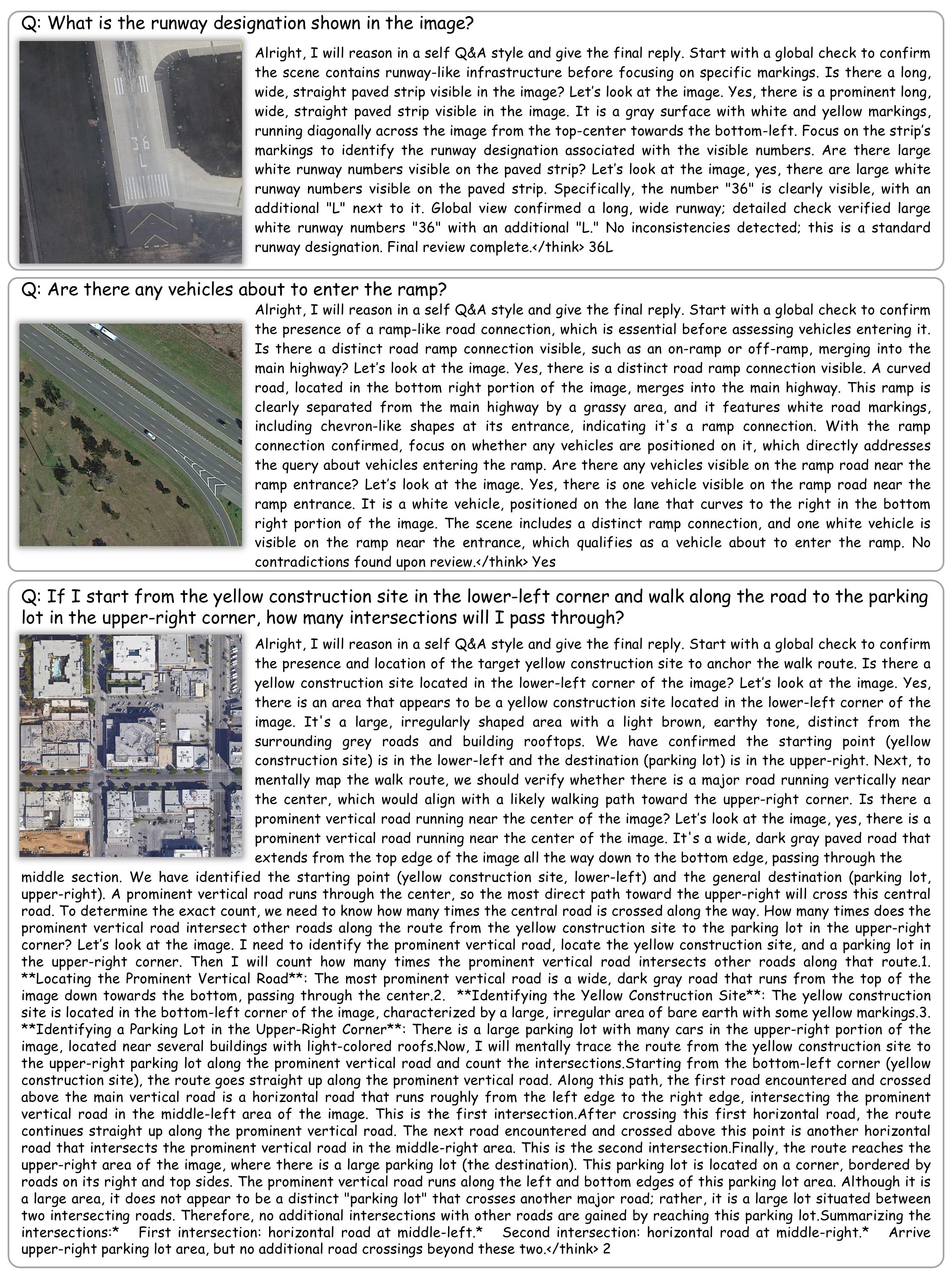}
  \vspace{-6pt}
  \caption{Reasoning cases of RS-EoT-7B (Part 1).}
  \label{fig:case_supp_1}
\end{figure*}

\begin{figure*}[t]
  \centering
  \includegraphics[width=0.95\linewidth]{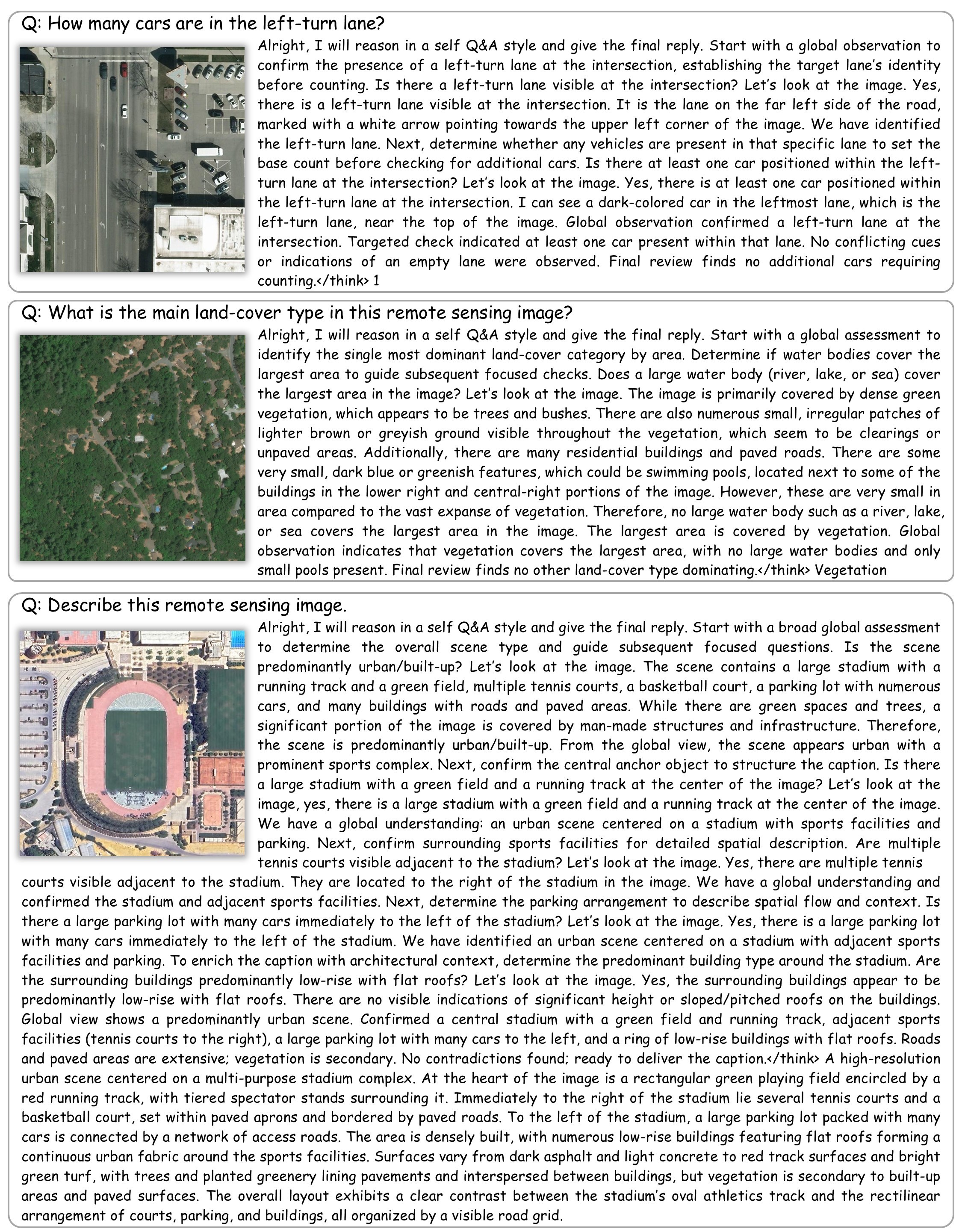}
  \vspace{-6pt}
  \caption{Reasoning cases of RS-EoT-7B (Part 2).}
  \label{fig:case_supp_2}
\end{figure*}

\begin{figure*}[t]
  \centering
  \includegraphics[width=0.95\linewidth]{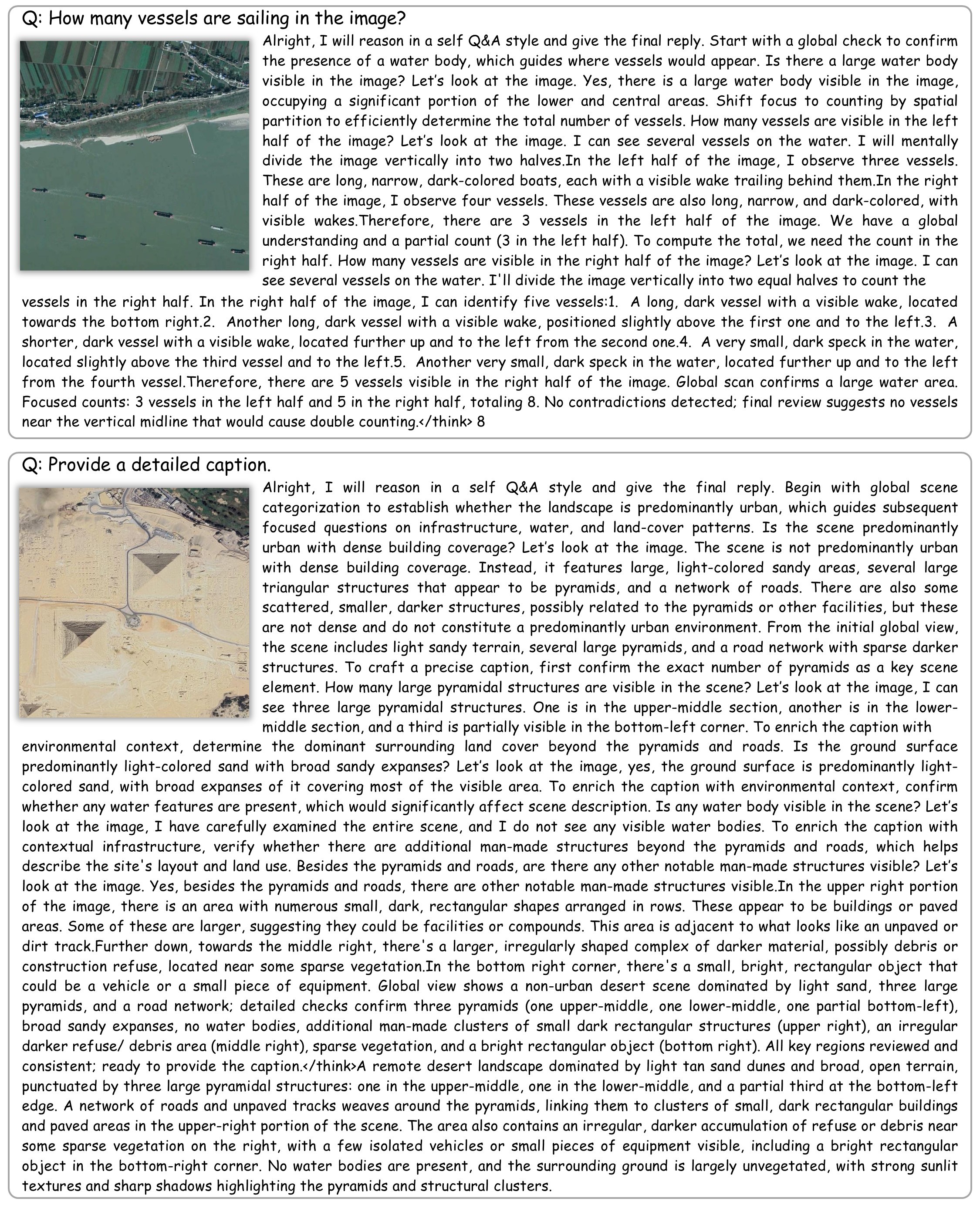}
  \vspace{-6pt}
  \caption{Reasoning cases of RS-EoT-7B (Part 3).}
  \label{fig:case_supp_3}
\end{figure*}

\begin{figure*}[t]
  \centering
  \includegraphics[width=0.95\linewidth]{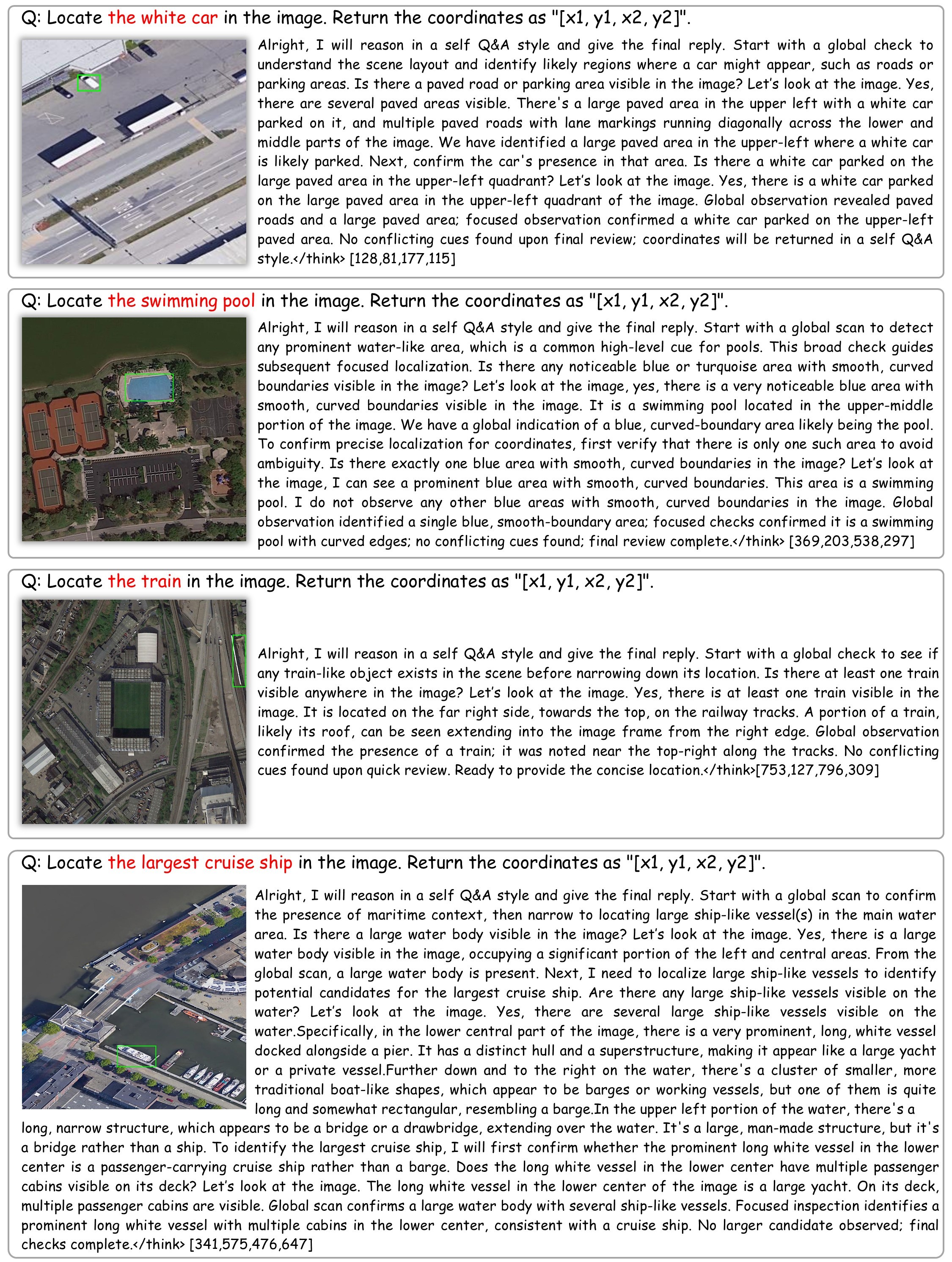}
  \vspace{-6pt}
  \caption{CReasoning cases of RS-EoT-7B (Part 4).}
  \label{fig:case_supp_4}
\end{figure*}

\begin{figure*}[t]
  \centering
  \includegraphics[width=0.95\linewidth]{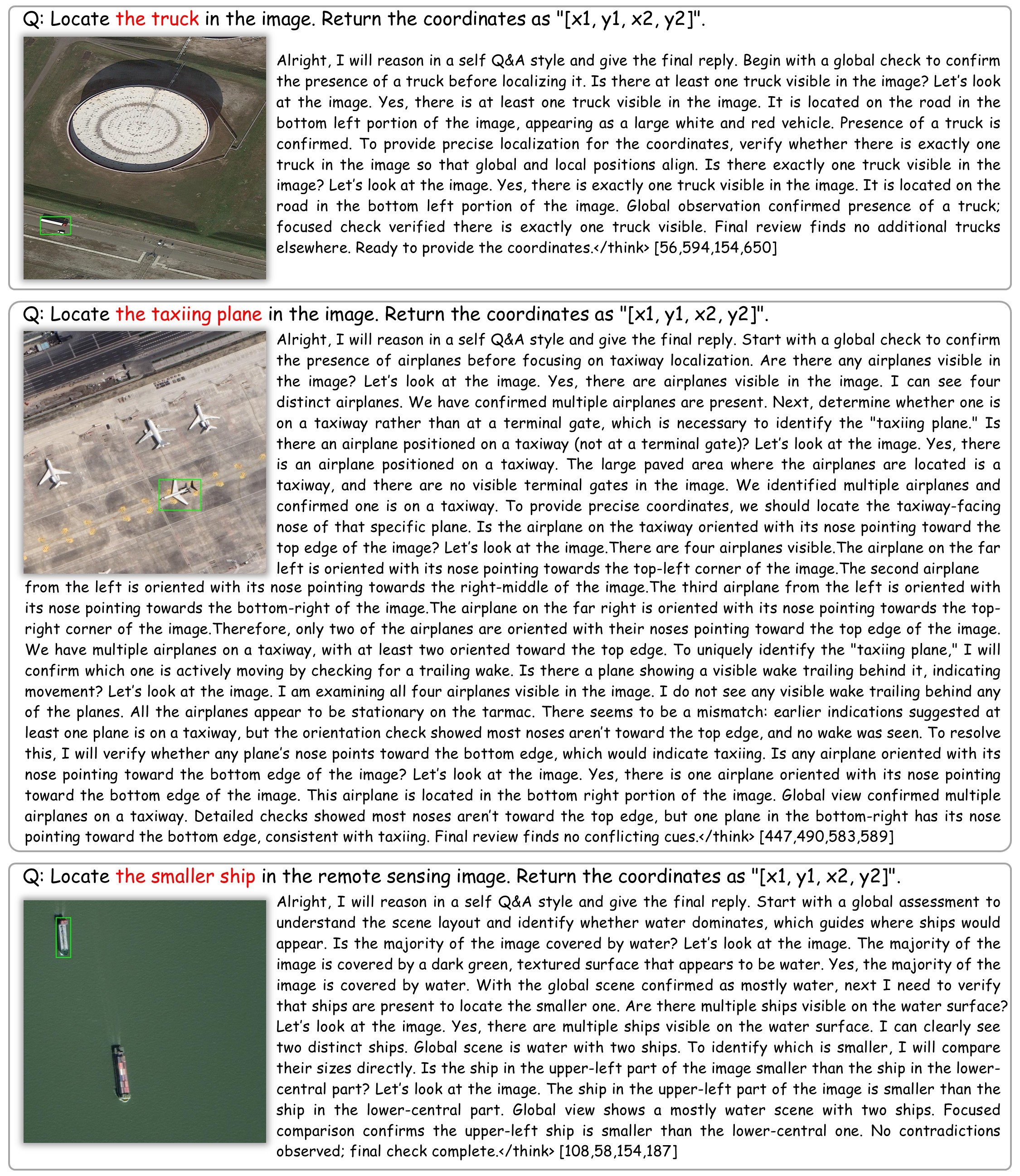}
  \vspace{-6pt}
  \caption{Reasoning cases of RS-EoT-7B (Part 5).}
  \label{fig:case_supp_5}
\end{figure*}

We provide additional qualitative examples to further demonstrate the effectiveness of RS-EoT-7B in complex remote sensing reasoning scenarios. Specifically, we present extended case studies covering both Remote Sensing General VQA tasks (\cref{fig:case_supp_1}, \cref{fig:case_supp_2}, and \cref{fig:case_supp_3}) and Fine-grained Grounding tasks (\cref{fig:case_supp_4} and \cref{fig:case_supp_5}). These visualizations illustrate how the model employs the iterative evidence-seeking paradigm to achieve robust reasoning and precise localization across diverse queries.

\end{document}